\documentclass[journal]{IEEEtran}
\IEEEpubidadjcol
\usepackage{amsmath,amsfonts}
\usepackage[hyperfootnotes=true]{hyperref}
\usepackage{algorithmic}
\usepackage{algorithm}
\usepackage{array}
\usepackage{textcomp}
\usepackage{stfloats}
\usepackage{url}
\usepackage{graphicx}
\usepackage{cite}
\usepackage{booktabs}
\usepackage{subfigure}
\usepackage{multirow}
\usepackage{threeparttable}
\usepackage{soul, color, xcolor}
\usepackage{hyperref}
\usepackage[normalem]{ulem}
\useunder{\uline}{\ul}{}
\begin{document}

\title{Ensemble Clustering via Co-association Matrix Self-enhancement}

\author{Yuheng~Jia,~\IEEEmembership{Member,~IEEE},~Sirui Tao,~Ran Wang,~\IEEEmembership{Member,~IEEE},~Yongheng~Wang
		\thanks{This work was supported in part by the National Natural Science Foundation of China under Grants 62106044, 62176160, in part by the Natural Science Foundation of Jiangsu Province under Grant BK20210221, in part by the National Key R\&D Program of China (Grant No. 2022YFF0608000), Zhejiang Lab (K2022KG1BB01), in part by the Guangdong Basic and Applied Basic Research Foundation (Grant 2022A1515010791), in part by the ZhiShan Youth Scholar Program from Southeast University 2242022R40015, and in part by the Natural Science Foundation of Shenzhen (Grant 20200804193857002). Corresponding author: Yongheng Wang.
		
		Y. Jia is with the School of Computer Science and Engineering, Southeast University, Nanjing 210096, China and also with the Research Center for Big Data Intelligence, Zhejiang Lab, Hangzhou 311121, China; 
        S. Tao is with the School of Automation, Southeast University, Nanjing 210096, China; 
        R. Wang is with the College of Mathematics and Statistics, Shenzhen University, Shenzhen 518060, China; 
        Y.H. Wang is with the Research Center for Big Data Intelligence, Zhejiang Lab, Hangzhou 311121, China. (e-mail: yhjia@seu.edu.cn, siruitao@seu.edu.cn, wangran@szu.edu.cn, wangyh@zhejianglab.com).}
	    }

\maketitle

\begin{abstract}
Ensemble clustering integrates a set of base clustering results to generate a stronger one.
Existing methods usually rely on a co-association (CA) matrix that measures how many times two samples are grouped into the same cluster according to the base clusterings to achieve ensemble clustering.
However, when the constructed CA matrix is of low quality, the performance will degrade.
In this paper, we propose a simple yet effective CA matrix self-enhancement framework that can improve the CA matrix to achieve better clustering performance.
Specifically, we first extract the high-confidence (HC) information from the base clusterings to form a sparse HC matrix. By propagating the highly-reliable information of the HC matrix to the CA matrix and complementing the HC matrix according to the CA matrix simultaneously, the proposed method generates an enhanced CA matrix for better clustering.
Technically, the proposed model is formulated as a symmetric constrained convex optimization problem, which is efficiently solved by an alternating iterative algorithm with convergence and global optimum theoretically guaranteed. Extensive experimental comparisons with twelve state-of-the-art methods on ten benchmark datasets substantiate the effectiveness, flexibility, and efficiency of the proposed model in ensemble clustering. The codes and datasets can be downloaded at \textcolor{red}{\url{https://github.com/Siritao/EC-CMS}}.
\end{abstract}

\begin{IEEEkeywords}
Ensemble clustering, Co-association matrix.
\end{IEEEkeywords}

\section{Introduction}
\IEEEPARstart{C}{lustering} is an important unsupervised machine learning task that divides samples into a set of groups to exploit their intrinsic patterns \cite{jain1999data}. In the past few years, many clustering methods have been proposed, like $K$-means \cite{hartigan1979algorithm,8767027}, mean-shift \cite{cheng1995mean}, hierarchical clustering \cite{jain1999data}, Gaussian mixture models \cite{bishop2006pattern}, spectral clustering \cite{von2007tutorial,8361074,8341858}, deep embedding clustering \cite{xie2016unsupervised}, and so on. 
Different clustering methods have their own benefits and drawbacks and usually consist of several hyper-parameters, which may fit different kinds of problems. However, how to select an appropriate clustering method for a specific problem and how to determine the hyper-parameters of the selected clustering method are still quite challenging. 
To this end, ensemble clustering (also known as consensus clustering) \cite{strehl2002cluster,GANAIE2022105151} was proposed to remedy this issue. 
Specifically, ensemble clustering first generates a set of clustering results by different clustering methods or by a typical clustering method with different hyper-parameters and initializations and then integrates those base clustering results to generate a consensus one, which is stronger than all the base clusterings.

\begin{figure}[t]
\centering
\includegraphics[width=3.3in]{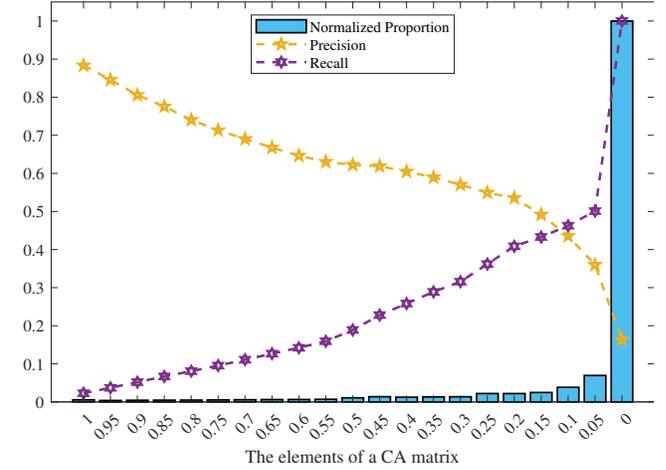}
\caption{First, $K$-means was used to generate 20 base clusterings on the Caltech20 dataset \cite{huang2017locally}, and then we produced a traditional CA matrix by \cite{fred2005combining} and recorded the frequencies of every two samples that are partitioned into the same cluster (the elements of the CA matrix).
We also calculated the precision and recall for different values in the CA matrix.
Note that we normalized the proportion of the elements of the CA matrix for better visualization.}
\label{pr}
\end{figure}

To achieve ensemble clustering, existing methods usually first construct a co-association (CA) matrix that records the co-occurrence relationships of samples, i.e., the frequencies of every pair of samples being grouped into the same cluster by the base clusterings \cite{fred2005combining}. As the CA matrix reflects the pairwise similarity among samples, it can serve as a similarity matrix or adjacency matrix. Then any typical similarity matrix-based clustering methods like spectral clustering or graph partitioning techniques \cite{strehl2002cluster,fern2004solving,li2012segmentation} can be applied to the CA matrix to generate the ensemble clustering result.
Recently, many efforts have been made to improve the quality of the traditional co-occurrence CA matrix based on different priors and assumptions. For example,
Li and Ding \cite{li2008weighted} globally weighted the CA matrix by ranking the quality of each base clustering result.
Huang \textit{et al.} \cite{huang2017locally} proposed the locally weighted CA (LWCA) matrix with an ensemble-driven cluster uncertainty estimation and the local weighting strategy.
In PTA \cite{huang2015robust}, the must-link relations among samples rather than all the individual links were focused on so they built a microcluster-based CA (MCA) matrix that is smaller in scale.
Some methods assume that the learned CA matrix holds a $K$-block-diagonal structure with $K$ being the number of clusters. For instance, Zhou \textit{et al.} \cite{zhou2020self} used Ky Fan's theorem \cite{fan1949theorem} to seek it and designed the CA matrix by self-paced learning, and further extended it to bipartite graph \cite{zhou2021self}.
Zhou \textit{et al.} \cite{zhou2021tri} tried to construct a robust CA matrix by multiple graph learning.
Apart from them, many approaches learned CA matrices with the low-rank assumption.
Yi \textit{et al.} \cite{yi2012robust} viewed reliable relations between samples as observed information and used the matrix completion technique to recover a low-rank CA matrix.
Zhou \textit{et al.} \cite{zhou2015learning} proposed to learn a robust and low-rank CA matrix by minimizing the Kullback–Leibler divergence.
Tao \textit{et al.} \cite{tao2019robust} learned a robust CA matrix with low-rank representation to seek the block-diagonal appearance of the CA matrix.
Although these methods can improve the clustering performance to some extent, if the used priors are not appropriate for specific tasks, the improvement will be degraded.

To further improve the clustering performance, in this paper we propose a CA matrix self-enhancement method to strengthen the quality of the CA matrix, without any extra information.
We found that in a CA matrix, a higher (resp. lower) value provides a more (resp. less) reliable description of the pairwise relationship between two samples, i.e., two samples are likely to be grouped into the same cluster,  while unfortunately, the majority of the elements in a CA matrix have low values. 
Fig.\,\ref{pr} vividly shows this observation. 
Motivated by this phenomenon, we first extract the high-confidence information from the base clustering results to constitute a sparse HC matrix, which provides limited but highly-reliable pairwise descriptions of samples.
Thereafter, we leverage the HC matrix and the original CA matrix to learn an ideal CA matrix. 
Specifically, for the HC matrix, we propose to propagate its information to the ideal CA matrix, while for the original CA matrix with many low-reliable but relatively dense connections, we try to denoise it to recover the ideal one.
By simultaneously using the information in the HC matrix and the original CA matrix, we are able to learn an enhanced CA matrix with relatively dense connections and more highly-reliable information. 
The proposed model is finally formulated as a Laplacian-regularized convex optimization problem, and we develop an efficient iterative optimization method to solve it with convergence and global optimum theoretically guaranteed. Extensive experiments on ten benchmark datasets validate its excellent clustering performance when compared with twelve recent ensemble clustering methods.
Moreover, the proposed model can adapt to multiple CA matrices and runs faster than most state-of-the-art methods with an iterative optimization process.
The main contributions of this paper are summarized as follows:
\begin{itemize}
\item We propose a novel ensemble clustering model from the perspective of CA matrix self-enhancement. The proposed model can promote different CA matrices without any extra information.
\item The proposed model is formulated as a convex optimization problem, which is solved by an alternating iterative algorithm with convergence guaranteed. It also runs faster than most of the state-of-the-art ensemble clustering methods with an iterative optimization process.
\item The proposed model exceeds the state-of-the-art ensemble clustering methods significantly  in clustering performance. Besides, the proposed model is robust to the hyper-parameters. 
\end{itemize}

The structure of this paper is organized as follows. We first review related works in Section \ref{sec2}, thereafter we introduce the proposed model with its numerical solution in Section \ref{sec3}. Section \ref{sec4} presents the experimental results and the associated analysis, and Section \ref{sec5} summarizes this paper.

\section{Related Work}
\label{sec2}
Given $n$ samples $\mathcal{X} = \{\mathbf{x}_1,...,\mathbf{x}_n\}$ and $m$ base clustering results $\boldsymbol{\Pi} = \{\boldsymbol{\pi}_1,...,\boldsymbol{\pi}_m\}$, where $\mathbf{x}_i$ denotes the $i$-th sample and $\boldsymbol{\pi}_i$ denotes the $i$-th base clustering that partitions $\mathcal{X}$ into several clusters $\mathcal{C}^i=\{\mathbf{c}_1^i,...,\mathbf{c}_{l_i}^i\}$ with $l_i$ being the number of clusters in $\mathcal{C}^i$.
According to Fred and Jain \cite{fred2005combining}, we can construct a co-occurrence CA matrix $\widetilde{\mathbf{A}}\in\mathbb{R}^{n\times n}$\footnote{Note that in this paper, we use $\widetilde{\mathbf{A}}$ to denote the traditional CA matrix in \cite{fred2005combining} that measures the frequency of two samples being grouped into the same cluster according to the base clusterings, and use $\mathbf{A}$ to denote more advanced (general) CA matrices (like LWCA) to be enhanced by our method.} by
\begin{equation}
\label{CA}
\widetilde{\mathbf{A}}_{i j}=\frac{1}{m} \sum_{k=1}^{m} \delta \left(\mathbf{c}^k\left(\mathbf{x}_i\right),\mathbf{c}^k\left(\mathbf{x}_j\right)\right),
\end{equation}
where $\mathbf{c}^k(\mathbf{x}_i) \in \mathcal{C}^k$ is the cluster membership of $\mathbf{x}_i$ in the $k$-th base clustering $\boldsymbol{\pi}_k$, and $\delta(\cdot,\cdot)$ is the Kronecker delta function:
\begin{equation*}
\delta(x,y)=\left\{
\begin{array}{ll}
1,   &   x=y\\
0,   &   x\neq y.\\
\end{array}
\right.
\end{equation*}
As $\widetilde{\mathbf{A}}_{ij}$ measures the frequency that $\mathbf{x}_i$ and $\mathbf{x}_j$ occur in the same cluster, it reveals the pairwise similarity between them according to the base clusterings $\boldsymbol{\Pi}$. Fred and Jain \cite{fred2005combining} apply hierarchical agglomerative clustering on the CA matrix to produce a consensus clustering result, which is known as evidence accumulation clustering (EAC). Hereafter, CA-based methods have become the mainstream of ensemble clustering due to their effectiveness and efficiency, and many advanced CA matrix construction methods were proposed \cite{li2008weighted,huang2017locally,huang2015robust,zhou2020self,zhou2021self,zhou2021tri,yi2012robust,zhou2015learning,tao2019robust}. In the following, we will introduce two representative ones.

\subsection{Locally Weighted Co-association}
Locally weighted ensemble clustering \cite{huang2017locally} refines the CA matrix by focusing on local diversity and cluster uncertainty. First, the uncertainty of a given cluster $\mathbf{c}_s^i$ from $\boldsymbol{\pi}_i$ is estimated by the entropy, i.e.,
\begin{equation*}
\mathrm{U}^{\boldsymbol{\Pi}}\left(\mathbf{c}_s^i\right) = -\sum_{j=1}^{m} \sum_{k=1}^{l_j} p\left(\mathbf{c}_s^i, \mathbf{c}_k^j\right) \log _{2} p\left(\mathbf{c}_s^i, \mathbf{c}_k^j\right).
\end{equation*}
Then the ensemble-driven cluster index (ECI) comes up to measure the reliability of a cluster that is negatively correlated with this entropy by a hyper-parameter $\theta$:
\begin{equation*}
\mathrm{ECI}\left(\mathbf{c}_s^i\right)=\mathrm{e}^{-\frac{\mathrm{U}^{\boldsymbol{\Pi}}\left(\mathbf{c}_s^i\right)}{\theta \cdot m}}.
\end{equation*}
Finally, the ECIs of all clusters are applied as local weights to compute a locally weighted CA (LWCA) matrix $\mathbf{A}\in\mathbb{R}^{n\times n}$:
\begin{equation*}
\mathbf{A}_{i j}=\frac{1}{m} \sum_{k=1}^{m} \delta \left(\mathbf{c}^k\left(\mathbf{x}_i\right),\mathbf{c}^k\left(\mathbf{x}_j\right)\right) \cdot \mathrm{ECI}\left(\mathbf{c}^k\left(\mathbf{x}_i\right)\right).
\end{equation*}

After building the LWCA matrix, hierarchical clustering or Tcut graph partitioning \cite{li2012segmentation} is used to generate the final partition.

\subsection{Microcluster Based Co-association and Probability Trajectory Similarity}
Probability trajectory accumulation (PTA) \cite{huang2015robust} first forms a compact microcluster-based CA (MCA) matrix. Specifically, if all the base clustering assigns a group of samples into one cluster, PTA will treat those samples indistinguishably, and aggregate them into a microcluster, which replaces those individual samples in the follow-up steps. PTA creates $n'$ non-overlapping microclusters $\mathcal{Y} = \{\mathbf{y}_1,...,\mathbf{y}_{n'}\}$, with each microcluster $\mathbf{y}_i$ containing $r_i$ samples. The MCA matrix $\mathbf{A}'\in\mathbb{R}^{n' \times n'}$ records the frequencies that every two microclusters occur in the same cluster from $m$ base clustering results $\boldsymbol{\Pi}$:
\begin{equation*}
\mathbf{A}'_{i j}=\frac{1}{m} \sum_{k=1}^{m} \delta \left(\mathbf{c}^k\left(\mathbf{y}_i\right),\mathbf{c}^k\left(\mathbf{y}_j\right)\right),
\end{equation*}
where $\mathbf{c}^k(\mathbf{y}_i) \in \mathcal{C}^k$ is the cluster that $\boldsymbol{\pi}_k$ assigns microcluster $\mathbf{y}_i$ to.

Afterward, PTA proposes an elite neighbor selection strategy to improve the quality of MCA. Specifically, a sparse similarity graph $\mathbf{W} =\{w_{ij}\}\in\mathbb{R}^{n'\times n'}$ is obtained by reserving the top-$V$ ($V$ is a hyper-parameter) links of each node and pruning other links, then the random walk is conducted on $\mathbf{W}$ to produce a dense similarity matrix. Considering that the transition probability from one node to another may be proportional to the size of microclusters and the weight of their link, the transition probability matrix $\mathbf{Q}=\{q_{ij}\}\in\mathbb{R}^{n'\times n'}$ can be calculated by:
\begin{equation*}
q_{i j}=\frac{r_j w_{ij}}{\sum_{i\neq k}r_k w_{ik}}.
\end{equation*}
Based on this, the $T$-step transition probability from $\mathbf{y}_i$ to $\mathbf{y}_j$ is represented by $\left[\mathbf{Q}^T\right]_{ij}$, and the probability trajectory of a random walker from node $\mathbf{y}_i$ with step length $T$ is defined as a $Tn'$-tuple $\mathbf{R}_i = \{\mathbf{Q}^1_{i,:},...,\mathbf{Q}^T_{i,:}\}$, where $\mathbf{Q}^T_{i,:}$ denotes the $i$-th row in $\mathbf{Q}^T$. The final probability trajectory similarity (PTS, also can be regarded as an advanced CA matrix) between two microclusters is the cosine similarity of their probability trajectory, i.e.,
\begin{equation*}
\mathbf{A}_{i j}=\frac{\langle \mathbf{R}_i,\mathbf{R}_j \rangle}{\sqrt{\langle \mathbf{R}_i,\mathbf{R}_i \rangle \cdot \langle \mathbf{R}_j,\mathbf{R}_j \rangle}},
\end{equation*}
where $\langle \cdot, \cdot \rangle$ denotes the inner product operation.

The PTS matrix serves as a similarity measure among microclusters. After applying hierarchical clustering or Tcut on it, the ensemble clustering result can be generated by mapping microclusters back to objects.

\textit{As the CA matrix plays a crucial role in ensemble clustering, in this work, we investigate how to improve its quality without additional information.}

\section{Proposed Method}
\label{sec3}
\subsection{Formulation}
In this section, we will present a novel CA matrix self-enhancement approach to promoting different kinds of CA matrices and accordingly produce better ensemble clustering performance. 
From Fig.\,\ref{pr} we can see that in a typical CA matrix, when the majority of base clusterings group two samples in the same cluster, those two samples are very likely to belong to the same class (with a high precision), but the total amount of this highly-reliable information is limited (with a low recall). 
On the contrary, there are a lot of low-value elements (with a high recall) in the CA matrix, which means the base clusterings provide relatively inconsistent descriptions (with low precision) of two samples.
Apparently, there exists a complementary relationship between those two kinds of information. As an ideal CA matrix is expected to be filled with dense and  highly-reliable elements, we, therefore, propose to leverage these two complementary pieces of information to enhance a given CA matrix $\mathbf{A}$\footnote{Note that $\mathbf{A}$ is a general similarity matrix that may have many concrete forms like the traditional co-occurrence CA matrix proposed by Fred  and Jain \cite{fred2005combining}, or LWCA \cite{huang2017locally}, PTS \cite{huang2015robust}, etc. These similarity matrices can all be enhanced via our model, see related results in Section \ref{adapt_CA}.}.

We first define a high-confidence (HC) matrix $\mathbf{H}=\{h_{ij}\} \in \mathbb{R}^{n \times n}$ to capture the limited but highly-reliable information from base clusterings $\boldsymbol{\Pi}$, i.e.,
\begin{equation}
\label{H}
\mathbf{H} = \Psi_\Omega(\widetilde{\mathbf{A}}),
\end{equation}
where $\Omega = \{(i,j) \mid \widetilde{\mathbf{A}}_{i j} \geq \alpha \}$ records the locations of the highly-reliable elements and $\Psi_\Omega: \mathbb{R}^{n \times n} \rightarrow \mathbb{R}^{n \times n}$ is a element-wise mapping operator:
\begin{equation*}
\left[ \Psi_{\Omega}(\widetilde{\mathbf{A}}) \right]_{i j}=\left\{
\begin{array}{ll}
\widetilde{\mathbf{A}}_{ij},   &   (i,j) \in \Omega\\
0,   &   (i,j)\notin \Omega.\\
\end{array}
\right.
\end{equation*}
Here $\widetilde{\mathbf{A}}$ is the traditional CA matrix described in Eq.\,(\ref{CA}), which directly captures the co-occurrence relationships among samples from base clustering results.
If the ratio of the times that two samples $\mathbf{x}_i$ and $\mathbf{x}_j$ are grouped into the same class to the total number of base clusterings exceeds a predefined threshold $\alpha\in[0,1]$, the corresponding entry $\widetilde{\mathbf{A}}_{ij}$ is viewed as a piece of highly-reliable information (as well as an element in the HC matrix). As the entries in $\widetilde{\mathbf{A}}$ are discrete and normalized in [0,1], while that in $\mathbf{A}$ usually has an arbitrary scale and distribution, we use $\widetilde{\mathbf{A}}$ rather than $\mathbf{A}$ to generate HC matrix, making it easy to determine the value of $\alpha$.

Second, we would directly use the given input CA matrix $\mathbf{A}$ to depict the relatively low-reliable information. Accordingly, \textit{CA matrix self-enhancement becomes using $\mathbf{H}$ and $\mathbf{A}$ to recover an ideal CA matrix $\mathbf{C} \in \mathbb{R}^{n \times n}$}.

As $\mathbf{A}$ may contain much inaccurate information, which can be regarded as the combination of the ideal CA matrix with noise. We therefore, try to remove the noise from $\mathbf{A}$ to construct $\mathbf{C}$, i.e.,

\begin{equation}
\begin{aligned} \label{initial}
&\min_{\mathbf{C},\,\mathbf{E}} \quad
\|\mathbf{E}\|_F^{2}\\
&\;\text{s.t.} \quad \mathbf{A}=\mathbf{C}+\mathbf{E},\, \mathbf{C}=\mathbf{C}^{\top},\,\mathbf{0} \leq \mathbf{C} \leq \mathbf{1},
\end{aligned}
\end{equation}
where $\mathbf{E}\in\mathbb{R}^{n\times n}$ denotes the error term and $\|\cdot\|_F$ denotes the Frobenius norm of a matrix. 
Ideally, $\mathbf{C}$ should be symmetric because if $\mathbf{x}_i$ is similar to $\mathbf{x}_j$, $\mathbf{x}_j$ should also be similar to $\mathbf{x}_i$.
Besides, with $\mathbf{C}_{ij}$ being the similarity measure, its value should lie in $[0,1]$.

However, solving Eq. (\ref{initial}) will lead to a trivial solution, i.e., $\mathbf{C}=\mathbf{A}$ and $\mathbf{E}=\mathbf{0}$. 
To avoid this, it is significant to further propagate the limited but highly valuable information from $\mathbf{H}$ to help build a reasonable $\mathbf{C}$.
Let $\mathbf{C}_{i,:}$ denote the $i$-th row of the matrix $\mathbf{C}$, which can reflect the pairwise similarity relationships of $\mathbf{x}_i$ with the other samples.
If $h_{ij}>0$, $\mathbf{x}_i$ and $\mathbf{x}_j$ are very likely to belong to the same cluster. Accordingly, the pairwise relationships of $\mathbf{x}_i$ with the other samples should also be close to that of $\mathbf{x}_j$, i.e., when $h_{ij}>0$, we can infer that $\mathbf{C}_{i,:}$ should be close to $\mathbf{C}_{j,:}$.
The intuition to minimize row-wise discrepancies with respect to every pair of samples that potentially belong to the same cluster can be mathematically expressed as:
\begin{equation*}
\min_{\mathbf{C}} \quad \sum_{i,j=1}^n h_{ij}\|\mathbf{C}_{i,:}-\mathbf{C}_{j,:}\|^2_2.
\end{equation*}
In addition, if $h_{ij}>0$, we can expect that $\mathbf{A}_{ij}$ is accurate enough to represent the pairwise similarity between $\mathbf{x}_i$ and $\mathbf{x}_j$. We thus copy those crucial entries of $\mathbf{A}$ to $\mathbf{C}$ directly, which would serve as the supporter of the co-association structure.
Taking all the above analyses into consideration, the proposed model is further formulated as:

\begin{equation}
\begin{aligned} \label{final}
&\min_{\mathbf{C},\,\mathbf{E}} \quad
\frac{1}{2} \sum_{i,j=1}^n h_{i j} \| \mathbf{C}_{i,:} - \mathbf{C}_{j,:} \|_2^2 + \frac{\lambda}{2} \|\mathbf{E}\|_F^{2}\\
&\;\text{s.t.} \quad \mathbf{A}=\mathbf{C}+\mathbf{E},\,\mathbf{C}_{i j}=\mathbf{A}_{i j},\,\forall(i,j) \in \Omega,\\
&\quad\quad\;\,\mathbf{C}=\mathbf{C}^{\top},\,\mathbf{0} \leq \mathbf{C} \leq \mathbf{1},
\end{aligned}
\end{equation}
where $\lambda>0$ balances the two loss terms. The first term can be written in a quadratic form $\operatorname{tr}(\mathbf{C}^{\top} \boldsymbol{\Phi} \mathbf{C})$ \cite{jia2021semisupervised}, where $\mathbf{\Phi}=\mathbf{D}-\mathbf{H}$ is the graph Laplacian matrix, and $\mathbf{D}$ is the diagonal degree matrix with its diagonal element $\mathbf{D}_{ii}=\sum_{j}\mathbf{H}_{ij}$. Also, operator $\Psi_{\Omega}$ can be  assigned to matrix $\mathbf{E}$ to compensate for unknown entries of $\mathbf{A}$ but locks $\mathbf{C}_{i j}=\mathbf{A}_{i j}$ for $(i,j) \in \Omega$. Finally, the proposed model is re-written  as:
\begin{equation}
\begin{aligned} \label{obj}
&\min_{\mathbf{C},\,\mathbf{E}} \quad
\operatorname{tr}\left(\mathbf{C}^{\top} \boldsymbol{\Phi} \mathbf{C}\right) + \frac{\lambda}{2}\|\mathbf{E}\|_F^{2}\\
&\; \text{s.t.} \quad \mathbf{A}=\mathbf{C}+\mathbf{E},\,
\Psi_\Omega(\mathbf{E})=\mathbf{0},\,\mathbf{C}=\mathbf{C}^{\top},\,\mathbf{0} \leq \mathbf{C} \leq \mathbf{1}.
\end{aligned}
\end{equation}

As a summary, Eq.\,(\ref{obj}) propagates the highly-reliable information in $\mathbf{H}$ and removes the noise in $\mathbf{A}$ simultaneously to learn an enhanced CA matrix $\mathbf{C}$ without any extra information. It can be applied to enhance different kinds of CA matrices (or similarity matrices) like $\widetilde{\mathbf{A}}$, LWCA, PTS, etc.
After solving Eq.\,(\ref{obj}), average-link hierarchical agglomerative clustering is applied on $\mathbf{C}$ to obtain the final clustering result. Other clustering approaches like spectral clustering are also applicable here.

\subsection{Optimization}

\begin{algorithm}[t]
\caption{Ensemble Clustering via Co-association Matrix Self-enhancement}
\label{algo}
\textbf{Input}: CA matrix $\mathbf{A}$, base clusterings $\boldsymbol{\Pi}$, threshold $\alpha$, trade-off parameter $\lambda$.\\
\textbf{Initialization}: Penalty parameter $\gamma_1 = \gamma_2 = 1$, tolerance $\epsilon = 1e-2$, $k=0$, $\mathbf{E}_k = \mathbf{C}_k = \mathbf{F}_k = \mathbf{0} \in \mathbb{R}^{n \times n}$, $\mathbf{Y}_{1k} = \mathbf{A} - \mathbf{C}_k - \mathbf{E}_k$, $\mathbf{Y}_{2k} = \mathbf{C}_k - \mathbf{F}_k$.

\begin{algorithmic}[1]
\STATE Construct the co-occurrence CA matrix $\widetilde{\mathbf{A}}$ by Eq.\,(\ref{CA}) from $\boldsymbol{\Pi}$, then generate the HC indices set $\Omega$ and derive the HC matrix $\mathbf{H}$ via Eq.\,(\ref{H}).
\STATE Compute $\boldsymbol{\Phi}=\operatorname{diag}[ \mathbf{H} \cdot(\mathbf{1}\in \mathbb{R}^{n})] - \mathbf{H}$.
\WHILE{not converged}
\STATE Update $\mathbf{C}_{k+1}$ via Eq.\,(\ref{Ck}).
\STATE Update $\mathbf{E}_{k+1}$ via Eq.\,(\ref{Ek}).
\STATE Update $\mathbf{F}_{k+1}$ via Eq.\,(\ref{Fk}).
\STATE Update $\mathbf{Y}_{1(k+1)}$, $\mathbf{Y}_{2(k+1)}$ via Eq.\,(\ref{Yk}).
\STATE Check stopping criteria:
\begin{equation*}
\mathop{\max}\left(\sigma_k(\mathbf{C}),\,\sigma_k(\mathbf{E}),\,\sigma_k(\mathbf{F}),\,\sigma_k(\mathbf{Y}_1),\,\sigma_k(\mathbf{Y}_2)\right)\leq \epsilon
\end{equation*}
with $\sigma_k(\mathbf{C}) = (\|\mathbf{C}_{k+1}-\mathbf{C}_k\|_F^{2}) / \|\mathbf{C}_k\|_F^{2}$.
\STATE $k \leftarrow k+1$.
\ENDWHILE
\STATE \textbf{return} the enhanced CA matrix $\mathbf{C}$.
\end{algorithmic}
\end{algorithm}

We adopt the Alternating Direction Method of Multipliers (ADMM) method \cite{boyd2011distributed} to solve Eq.\,(\ref{obj}), which can handle the equality constraint and the multiple variables effectively. In order to alleviate the value range constraint and the symmetric constraint of $\mathbf{C}$, we introduce an intermediate matrix $\mathbf{F}$ to shoulder them and equivalently re-write Eq.\,(\ref{obj}) as 
\begin{equation}
\begin{aligned} \label{final_obj}
&\min_{\mathbf{C},\,\mathbf{E},\,\mathbf{F}} \quad
\operatorname{tr}\left(\mathbf{C}^{\top} \boldsymbol{\Phi} \mathbf{C}\right) + \frac{\lambda}{2}\|\mathbf{E}\|_F^{2}\\
&\; \text{s.t.} \quad \mathbf{A}=\mathbf{C}+\mathbf{E},\,
\Psi_\Omega(\mathbf{E})=\mathbf{0},\,\mathbf{C}=\mathbf{F},\\
&\quad \quad\;\;\mathbf{F}=\mathbf{F}^{\top},\,\mathbf{0} \leq \mathbf{F} \leq \mathbf{1}.
\end{aligned}
\end{equation}
Let $\mathbf{Y}_1$ and $\mathbf{Y}_2$ denote the Lagrangian multipliers, the augmented Lagrangian function of Eq.\,(\ref{final_obj}) can be written as:
\begin{equation}
\begin{aligned} \label{Lagrangian}
&\mathcal{L}\left( \mathbf{C},\,\mathbf{E},\,\mathbf{F},\,\mathbf{Y}_1,\,\mathbf{Y}_2 \right) = \operatorname{tr}\left(\mathbf{C}^{\top} \boldsymbol{\Phi} \mathbf{C} \right)+\frac{\lambda}{2}\|\mathbf{E}\|_F^{2}\\
&+ \langle \mathbf{Y}_1, \mathbf{A}-\mathbf{C}-\mathbf{E} \rangle + \frac{\gamma_1}{2}\|\mathbf{A}-\mathbf{C}-\mathbf{E}\|_F^{2}\\
&+ \langle \mathbf{Y}_2, \mathbf{C}-\mathbf{F} \rangle + \frac{\gamma_2}{2}\|\mathbf{C}-\mathbf{F}\|_F^{2}\\
&\; \text{s.t.} \quad \Psi_\Omega(\mathbf{E})=\mathbf{0},\,\mathbf{F}=\mathbf{F}^{\top},\,\mathbf{0} \leq \mathbf{F} \leq \mathbf{1},
\end{aligned}
\end{equation}
where $\gamma_1,\gamma_2>0$ introduce the augmented Lagrangian terms. In the experiments, we fixed $\gamma_1=\gamma_2=1$, which is a good choice for many ADMM-based optimization solvers \cite{boyd2011distributed}.
The solution to Eq.\,(\ref{obj}) can be obtained when Eq.\,(\ref{Lagrangian}) reaches its minimum. Specifically, Eq.\,(\ref{Lagrangian}) can be solved by the following iterative alternating procedures.

\subsubsection{Updating \textbf{C}} With other variables fixed, the $\mathbf{C}$ sub-problem is formulated as:
\begin{equation}
\begin{aligned} \label{C1}
&\mathbf{C}_{k+1} = \mathop{\arg\min}_{\mathbf{C}} \operatorname{tr}\left(\mathbf{C}^{\top} \boldsymbol{\Phi} \mathbf{C}\right)+\frac{\gamma_1}{2}\|\mathbf{C}- \mathbf{P}_{1k}\|_F^{2}\\
&\quad\quad\;\,+\frac{\gamma_2}{2}\|\mathbf{C}- \mathbf{P}_{2k}\|_F^{2},
\end{aligned}
\end{equation}
where $\mathbf{P}_{1k} = \mathbf{A} -\mathbf{E}_k + \mathbf{Y}_{1k}/\gamma_1$ and $\mathbf{P}_{2k} = \mathbf{F}_k - \mathbf{Y}_{2k}/\gamma_2$.
By setting the derivative of Eq.\,(\ref{C1}) to $0$, we have the following analytical solution of the $\mathbf{C}$ sub-problem:
\begin{equation}
\label{Ck}
\mathbf{C}_{k+1} = \left(2\boldsymbol{\Phi} +\left(\gamma_1 + \gamma_2 \right) \mathbf{I} \right)^{-1} \left( \gamma_1 \mathbf{P}_{1k} +\gamma_2 \mathbf{P}_{2k} \right).
\end{equation}

\subsubsection{Updating \textbf{E}} With other variables fixed, the formulation of the $\mathbf{E}$ sub-problem is:
\begin{equation}
\begin{aligned} \label{E}
&\mathbf{E}_{k+1} = \mathop{\arg\min}_{\mathbf{E}} \frac{\gamma_1}{2}\|\mathbf{E}-\left(\mathbf{A}-\mathbf{C}_{k+1}\right)\|_F^{2} +\frac{\lambda}{2}\|\mathbf{E}\|_F^{2}\\
&\; \quad \quad + \langle \mathbf{Y}_{1k}, \mathbf{A}-\mathbf{C}_{k+1}-\mathbf{E} \rangle \\
&\;\text{s.t.} \quad \Psi_\Omega(\mathbf{E})=\mathbf{0}.
\end{aligned}
\end{equation}
As the Frobenius norm, the inner product, and the equality constraint are all computed element-wisely, we can get the global solution of Eq.\,(\ref{E}) through:
\begin{equation}
\begin{aligned}
\label{Ek}
&\mathbf{E}^* = \frac{\gamma_1 \left(\mathbf{A}-\mathbf{C}_{k+1} \right) +\mathbf{Y}_{1k}}{\lambda+\gamma_1},
&\mathbf{E}_{k+1}=\Psi_{\bar{\Omega}}(\mathbf{E}^*),
\end{aligned}
\end{equation}
where $\bar{\Omega}$ is the complementary set of $\Omega$.

\subsubsection{Updating \textbf{F}} With other variables fixed, the $\mathbf{F}$ sub-problem can be expressed as:
\begin{equation}
\begin{aligned}
\label{F}
&\mathbf{F}_{k+1} = \mathop{\arg\min}_{\mathbf{F}} \frac{1}{2} \|\mathbf{F}-\mathbf{P}_{3k}\|_F^{2}\\
&\;\text{s.t.} \quad \mathbf{F}=\mathbf{F}^{\top},\,\mathbf{0} \leq \mathbf{F} \leq \mathbf{1},
\end{aligned}
\end{equation}
where $\mathbf{P}_{3k} = \mathbf{C}_{k+1} +\mathbf{Y}_{2k}/\gamma_2$.
Since $\mathbf{F}=\mathbf{F}^{\top}$ suggests that $\|\mathbf{F}-\mathbf{P}_{3k}\|_F^{2}=\|\mathbf{F}-\mathbf{P}_{3k}^\top\|_F^{2}$, Eq.\,(\ref{F}) is equivalent to:
\begin{equation*}
\begin{aligned}
&\mathbf{F}_{k+1} = \mathop{\arg\min}_{\mathbf{F}} \frac{1}{4} \left( \|\mathbf{F}-\mathbf{P}_{3k}\|_F^{2}+\|\mathbf{F}-\mathbf{P}_{3k}^\top\|_F^{2} \right)\\
&\; \quad \quad = \mathop{\arg\min}_{\mathbf{F}} \frac{1}{2} \left\|\mathbf{F}-\frac{\mathbf{P}_{3k} + \mathbf{P}_{3k}^\top}{2}\right\|_F^{2} + c\left(\mathbf{P}\right)\\
&\;\text{s.t.} \quad \mathbf{F}=\mathbf{F}^{\top},\,\mathbf{0} \leq \mathbf{F} \leq \mathbf{1}.
\end{aligned}
\end{equation*}
with $c\left(\mathbf{P}\right)$ irrelevant to $\mathbf{F}$. Thereafter, the optimum of Eq.\,(\ref{F}) is achieved through an element-wisely truncation:
\begin{equation}
\label{Fk}
\mathbf{F}_{k+1}=\mathop{\min}\left( \mathop{\max}\left( \frac{\mathbf{P}_{3k} + \mathbf{P}_{3k}^\top}{2},0\right),1 \right).
\end{equation}

\subsubsection{Updating \textbf{Y}}
The ADMM algorithm updates the multiplier matrices \cite{7410679} $\mathbf{Y}_1,\,\mathbf{Y}_2$ by:
\begin{equation}
\label{Yk}
\begin{aligned}
&\mathbf{Y}_{1(k+1)}=\mathbf{Y}_{1k}+\gamma_1 \left(\mathbf{A}- \mathbf{C}_{k+1}- \mathbf{E}_{k+1} \right), \\
&\mathbf{Y}_{2(k+1)}=\mathbf{Y}_{2k}+\gamma_2 \left(\mathbf{C}_{k+1}- \mathbf{F}_{k+1} \right).
\end{aligned}
\end{equation}

\begin{table}[t]
\centering
\caption{Description of the datasets with a modest scale.}
\label{tab:dataset}
\begin{threeparttable}
\begin{tabular}{@{}ccccc@{}}
\toprule
Dataset     & \#Instance & \#Feature & \#Class & Source                                 \\ \midrule
Caltech20 \tnote{1}   & 2,386      & 30,000    & 20      & IMAGE, \cite{fei2004learning,huang2017locally}  \\
Ecoli \tnote{2}      & 336        & 8         & 8       & UCI                                      \\
LS \tnote{3}         & 6,435      & 36        & 6       & UCI                                      \\
FCT \tnote{4}        & 3,780      & 54        & 7       & UCI                                      \\
Aggregation \tnote{5} & 788        & 2         & 7       & SHAPE,\cite{10.1145/1217299.1217303}                                      \\
Texture \tnote{6}    & 5,500      & 40        & 11      & UCI                                      \\
UMIST \tnote{7}      & 575        & 644       & 20      & FACE, \cite{wechsler2012face}
          \\
SPF \tnote{8}        & 1,941      & 27        & 7       & UCI                                      \\ \bottomrule
\end{tabular}

\begin{tablenotes}
       \footnotesize
       \item[1] \url{https://data.caltech.edu/records/mzrjq-6wc02}
       \item[2] \url{https://archive.ics.uci.edu/ml/datasets/ecoli}
       \item[3] \url{https://archive.ics.uci.edu/ml/datasets/Statlog+(Landsat+Satellite)}
       \item[4] \url{https://archive.ics.uci.edu/ml/datasets/covertype}
       \item[5] \url{http://cs.uef.fi/sipu/datasets/}
       \item[6] \url{https://www.elen.ucl.ac.be/neural-nets/Research/Projects/ELENA/databases/REAL/texture/}
       \item[7] \url{http://images.ee.umist.ac.uk/danny/database.html}
       \item[8] \url{https://archive.ics.uci.edu/ml/datasets/steel+plates+faults}
     \end{tablenotes}
     
\end{threeparttable}
\end{table}

\begin{table}[t]
\centering
\caption{Description of two large-scale datasets.}
\label{tab:large-dataset}
\begin{threeparttable}
\begin{tabular}{@{}ccccc@{}}
\toprule
Dataset     & \#Instance & \#Feature & \#Class & Source                                 \\ \midrule
ISOLET \tnote{1}   & 7,797       & 617         & 26       & UCI                                      \\
USPS \tnote{2}     & 11,000      & 256         & 10       & IMAGE                                      \\ \bottomrule
\end{tabular}

\begin{tablenotes}
       \footnotesize
       \item[1] \url{https://archive.ics.uci.edu/ml/datasets/isolet}
       \item[2] \url{https://cs.nyu.edu/~roweis/data.html}
     \end{tablenotes}

\end{threeparttable}
\end{table}

\begin{figure*}[t]
\centering
\subfigure{
\begin{minipage}[t]{0.238\linewidth}
\includegraphics[width=1.7in]{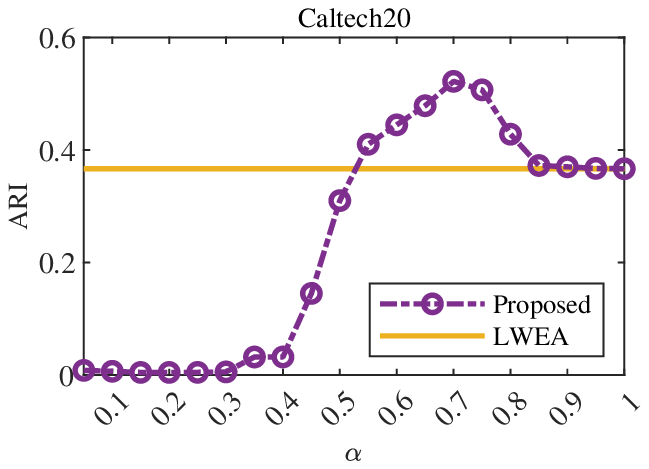}
\end{minipage}}
\subfigure{
\begin{minipage}[t]{0.238\linewidth}
\includegraphics[width=1.7in]{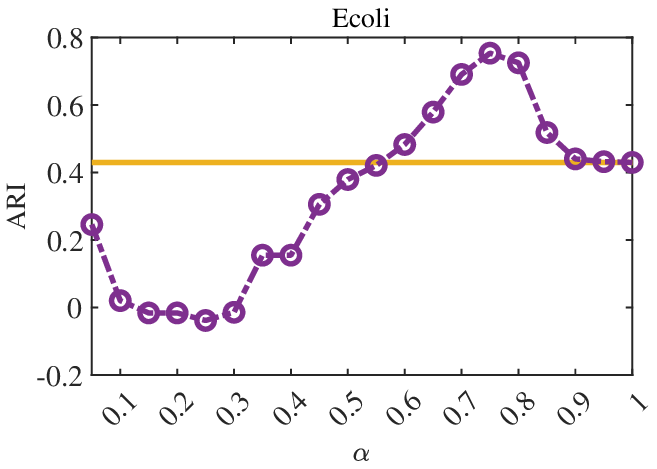}
\end{minipage}}
\subfigure{
\begin{minipage}[t]{0.238\linewidth}
\includegraphics[width=1.7in]{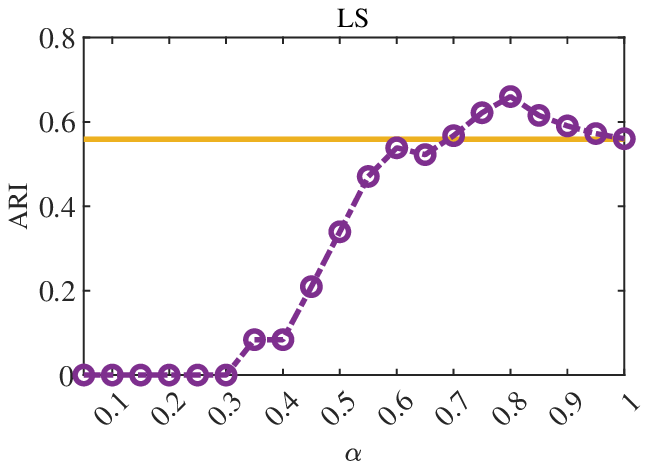}
\end{minipage}}
\subfigure{
\begin{minipage}[t]{0.238\linewidth}
\includegraphics[width=1.7in]{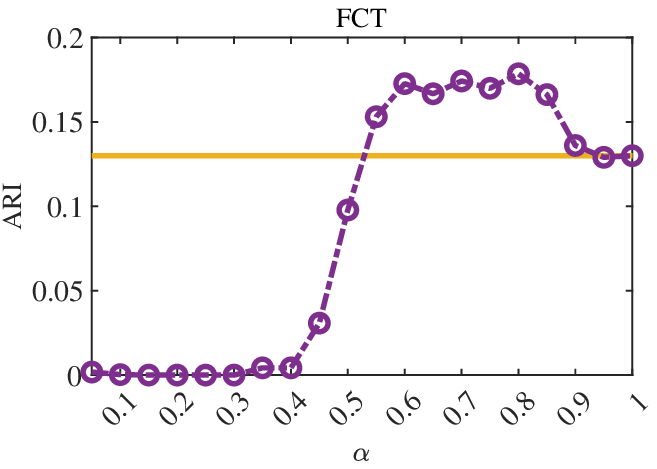}
\end{minipage}}
\subfigure{
\begin{minipage}[t]{0.238\linewidth}
\includegraphics[width=1.7in]{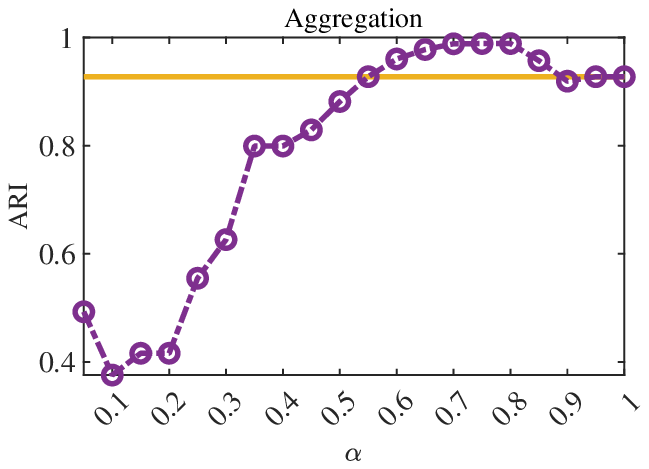}
\end{minipage}}
\subfigure{
\begin{minipage}[t]{0.238\linewidth}
\includegraphics[width=1.7in]{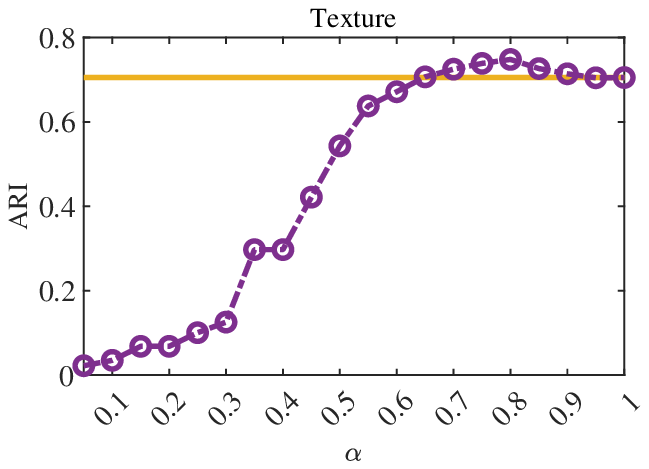}
\end{minipage}}
\subfigure{
\begin{minipage}[t]{0.238\linewidth}
\includegraphics[width=1.7in]{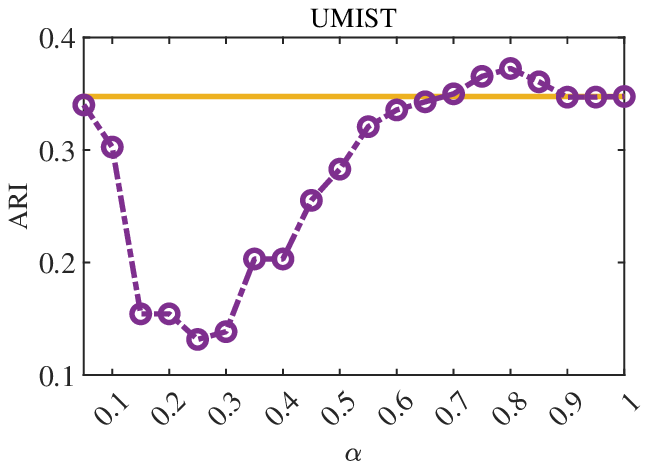}
\end{minipage}}
\subfigure{
\begin{minipage}[t]{0.238\linewidth}
\includegraphics[width=1.7in]{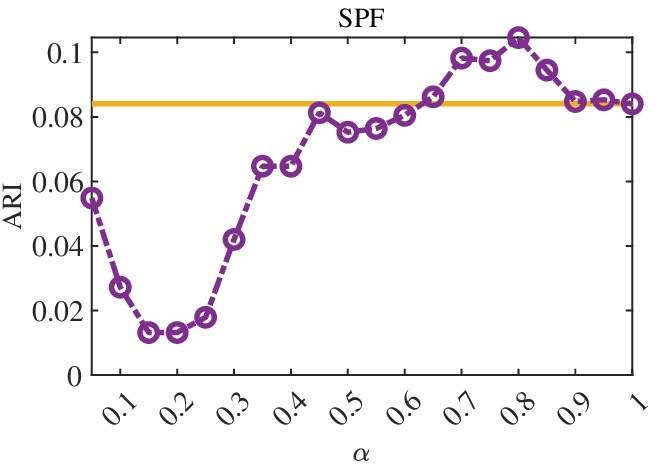}
\end{minipage}}
\caption{Clustering performances w.r.t. ARI with varying $\alpha$. Both LWEA and the proposed method adopt average-link hierarchical agglomerative clustering on the CA matrix to produce the clustering result. All the sub-figures share the same legend.}
\label{varyalp}
\end{figure*}

\begin{figure*}[t]
\centering
\subfigure{
\begin{minipage}[t]{0.238\linewidth}
\includegraphics[width=1.7in]{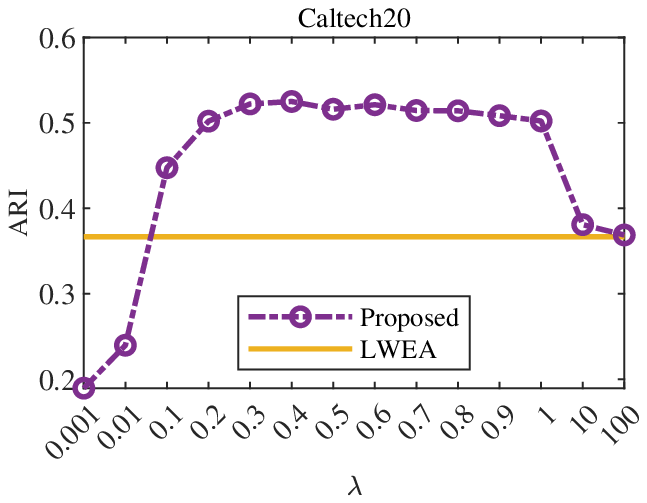}
\end{minipage}}
\subfigure{
\begin{minipage}[t]{0.238\linewidth}
\includegraphics[width=1.7in]{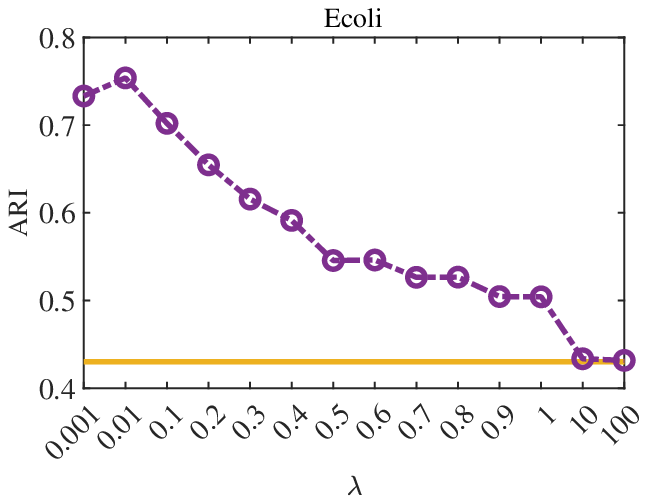}
\end{minipage}}
\subfigure{
\begin{minipage}[t]{0.238\linewidth}
\includegraphics[width=1.7in]{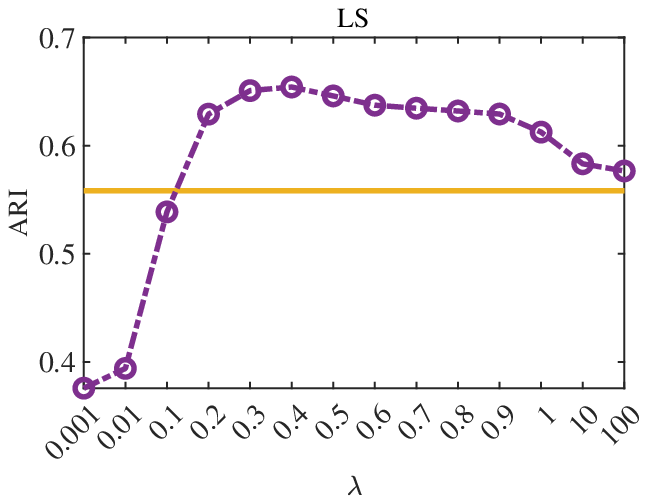}
\end{minipage}}
\subfigure{
\begin{minipage}[t]{0.238\linewidth}
\includegraphics[width=1.7in]{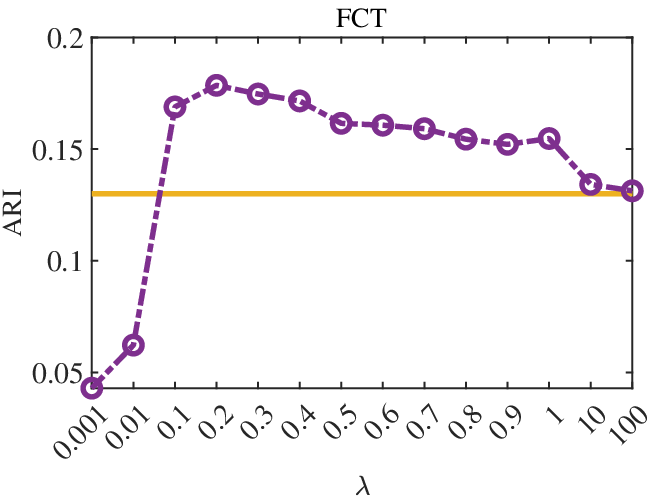}
\end{minipage}}
\subfigure{
\begin{minipage}[t]{0.238\linewidth}
\includegraphics[width=1.7in]{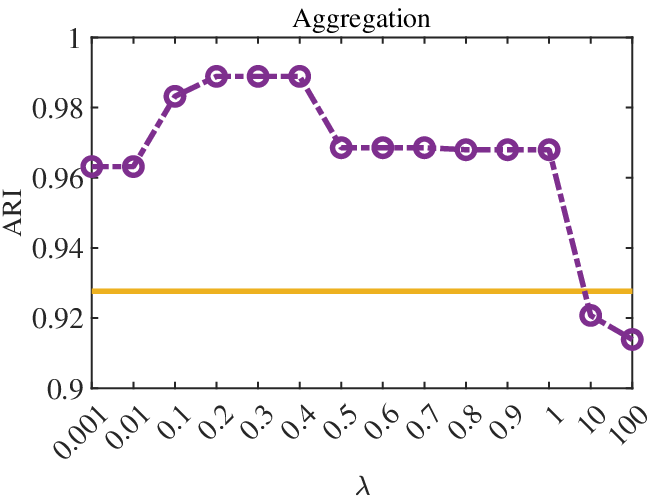}
\end{minipage}}
\subfigure{
\begin{minipage}[t]{0.238\linewidth}
\includegraphics[width=1.7in]{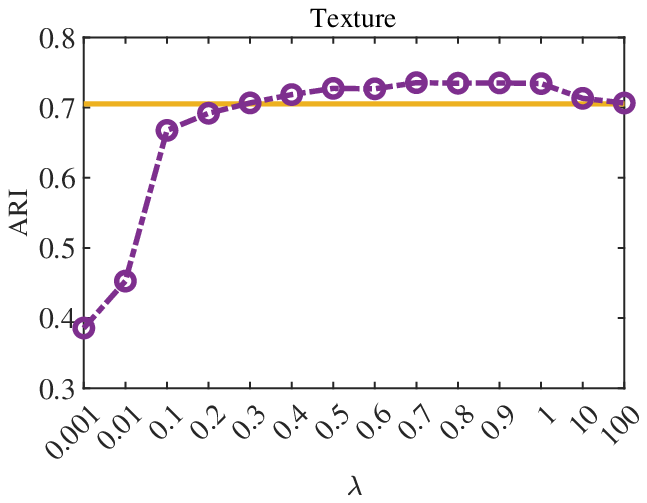}
\end{minipage}}
\subfigure{
\begin{minipage}[t]{0.238\linewidth}
\includegraphics[width=1.7in]{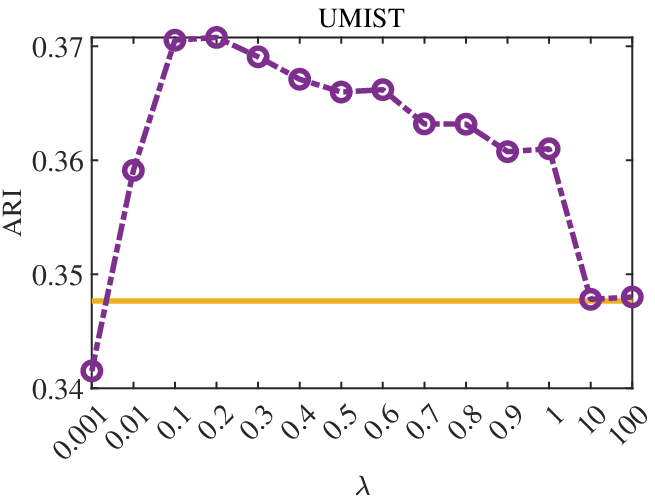}
\end{minipage}}
\subfigure{
\begin{minipage}[t]{0.238\linewidth}
\includegraphics[width=1.7in]{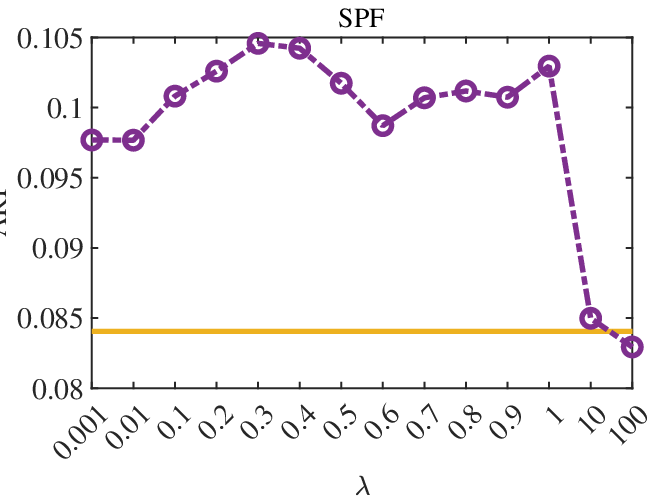}
\end{minipage}}
\caption{Clustering performances w.r.t. ARI with varying $\lambda$. Both LWEA and the proposed method adopt average-link hierarchical agglomerative clustering on the CA matrix to produce the clustering result. All the sub-figures share the same legend.}
\label{varylam}
\end{figure*}

\begin{table*}[]
\centering
\caption{Influence of ADMM parameters ($\gamma_1$ and $\gamma_2$) on the UMIST dataset ($\alpha$ and $\lambda$ were set to the recommended value). We fixed one of them and varied the other, then recorded the offset of ARI, NMI, and F-score compared with default setting ($\gamma_1=\gamma_1=1$), and the average variation ratio of $\mathbf{C}$, measured by Frobenius norm.}
\label{tab:ADMM_para}
\begin{tabular}{@{}c|ccccc|ccccc@{}}
\toprule
ADMM Parameter             & \multicolumn{5}{c|}{$\gamma_1=1$, $\gamma_2=$} & \multicolumn{5}{c}{$\gamma_2=1$, $\gamma_1=$}      \\ \midrule
Measure                    & 0.01   & 0.1  & 1              & 10   & 100    & 0.01   & 0.1    & 1              & 10     & 100    \\ \midrule
ARI                        & +0.001 & 0    & \textbf{0.367} & 0    & 0      & -0.034 & -0.001 & \textbf{0.367} & -0.002 & -0.012 \\
NMI                        & +0.001 & 0    & \textbf{0.682} & 0    & +0.003 & -0.035 & 0      & \textbf{0.682} & -0.001 & -0.010 \\
F-score                    & +0.001 & 0    & \textbf{0.406} & 0    & 0      & -0.028 & 0      & \textbf{0.406} & -0.002 & -0.013 \\ \midrule
$ \frac{\lvert\|\mathbf{C}_{\gamma_1,\gamma_2}\|_F-\|\mathbf{C}_{\gamma_1=\gamma_2=1}\|_F\rvert}{\|\mathbf{C}_{\gamma_1=\gamma_2=1}\|_F}$ & 0.05   & 0.02 & 0              & 0.02 & 0.38   & 0.38   & 0.05   & 0              & 0.03   & 0.11   \\ \bottomrule
\end{tabular}
\end{table*}

The whole procedure is summarized in Algorithm\,\ref{algo}. As Eq.\,(\ref{obj}) is a convex problem with two blocks of variables (i.e., $\{\mathbf{C}\}$ and $\{\mathbf{E}, \mathbf{F}\}$, since updating variables $\mathbf{E}$ and $\mathbf{F}$ are independent of each other), Algorithm\,\ref{algo} is theoretically guaranteed to converge to a global solution by ADMM \cite{boyd2011distributed}.

\subsection{Computational Complexity Analysis}
Updating $\mathbf{C}$ involves a matrix inverse operation and a matrix multiplication operation. As in each iteration, the to-be-inversed matrix $2\mathbf{\Phi}+(\gamma_1+\gamma_2)\mathbf{I}$ is fixed so that can be calculated in advance, which means the computational complexity of solving the $\mathbf{C}$ sub-problem is $\mathcal{O}(n^{2.37})$ \cite{alman2021refined}.
Solving sub-problems of $\mathbf{E}$, $\mathbf{F}$ and updating multiplier matrices $\mathbf{Y}_1$, $\mathbf{Y}_2$ only include element-wise calculations, leading to computational complexity of $\mathcal{O}(n^{2})$. 
Therefore, the overall computational complexity of Algorithm\,\ref{algo} is $\mathcal{O}(n^{2.37})$ in one iteration.

\section{Experiment}
\label{sec4}
We conducted a series of experiments on eight real-world benchmark datasets with a modest scale, and two large-scale datasets to verify the effectiveness of the proposed algorithm, whose detailed information is summarized in Tables\,\ref{tab:dataset} and \ref{tab:large-dataset}. Following Huang \textit{et al.} \cite{huang2017locally}, for all the datasets, 100 candidate base clusterings were generated by $K$-means with $K$ randomly selecting in $[2, \sqrt{n}]$ and $n$ being the number of samples.

We compared the proposed method with twelve representative state-of-the-art ensemble clustering algorithms, which are listed as follows.
\begin{enumerate}
    \item PTA-AL, PTA-CL, PTGP \cite{huang2015robust}: three microcluster representation-based ensemble clustering methods using similarity matrix derived from probability trajectories of random walkers. They adopt average-link (AL) hierarchical agglomerative clustering, complete-link (CL) hierarchical agglomerative clustering and Tcut graph partitioning (GP) \cite{li2012segmentation} respectively to generate clustering results.
    \item LWEA, LWGP \cite{huang2017locally}: Two locally-weighted ensemble clustering methods with average-link hierarchical agglomerative clustering and Tcut, corresponding to LWEA (evidence accumulation) and LWGP (graph partitioning). The local weights are obtained via an ensemble-driven cluster uncertainty estimation.
    \item RSEC-H, RSEC-Z \cite{tao2019robust}: robust spectral ensemble clustering methods via rank minimization. The algorithm simultaneously learns a consensus partition $\mathbf{H}$ and a low-rank representation $\mathbf{Z}$, and respectively adopts $K$-means or spectral clustering on them to get final results.
    \item DREC \cite{zhou2019ensemble}: a dense representation-based ensemble clustering algorithm. It also uses the idea of forming microclusters but with the consideration of outliers.
    \item SPCE \cite{zhou2020self}: ensemble clustering with a self-paced manner. It learns a CA matrix from easy-to-learn samples to difficult-to-learn ones.
    \item ECPCS-MC, ECPCS-HC \cite{8525437}: two ensemble clustering methods that improve the CA matrix by propagating cluster-wise similarities. Here, MC refers to cluster-level meta-clustering and HC refers to average-link hierarchical agglomerative clustering.
    \item TRCE \cite{zhou2021tri}: a multiple graph learning-based ensemble clustering method combining three levels of robustness, i.e., base clustering level, graph level, and instance level.
\end{enumerate}
For a fair comparison, we carefully tuned the hyper-parameters of those compared methods according to their original papers and reported the best performances.
As for the proposed method, the LWCA matrix \cite{huang2017locally} was employed as the input CA matrix.
For all the methods, we randomly selected 20 base clusterings from the candidate base clustering pool and recorded the average clustering performance with the  standard deviation over 20 repetitions. All experiments were conducted by MATLAB R2020a on a PC with a 2.3GHz CPU and 16GB memory.

We adopted five widely-used metrics, i.e., adjusted rand index (ARI) \cite{vinh2010information}, normalized mutual information (NMI) \cite{strehl2002cluster}, F-score (the harmonic mean of precision and recall), Accuracy and Purity to evaluate the clustering performance of different methods. All of them lie in the range of $[0,1]$, and larger values reflect better clustering performance.

\subsection{Parameter Sensitivity}

\begin{table*}[t]
\centering
\caption{Clustering performances of different algorithms measured by ARI, where the best performance and the second best one are highlighted by bold and underline.}
\label{tab:ari}
\renewcommand{\arraystretch}{1.05}
\resizebox{\textwidth}{!}{
\begin{tabular}{@{}lcccccccc@{}}
\toprule
\multicolumn{1}{c}{Method}     & Caltech20            & Ecoli                & LS                   & FCT                  & Aggregation          & Texture              & UMIST                & SPF                  \\ \midrule
Base clusterings (average)       & 0.209±0.061                & 0.396±0.120                & 0.235±0.100                & 0.075±0.017                & 0.463±0.171                & 0.373±0.088                & 0.273±0.0.092                & 0.044±0.013                \\
Base clusterings (best)       & 0.345                & 0.695                & 0.479                & 0.120                & 0.822                & 0.530                & 0.389                & 0.073                \\ \midrule
PTA-AL {[}\textit{TKDE}, 2016{]}         & 0.356±0.034          & 0.340±0.043          & 0.593±0.079    & 0.136±0.017          & 0.600±0.128          & 0.658±0.046          & 0.323±0.027          & 0.081±0.015          \\
PTA-CL {[}\textit{TKDE}, 2016{]}         & 0.334±0.026          & 0.342±0.045          & 0.452±0.068          & 0.124±0.015          & 0.529±0.181          & 0.604±0.027          & 0.323±0.029          & 0.075±0.018          \\
PTGP {[}\textit{TKDE}, 2016{]}           & 0.335±0.042          & 0.351±0.038          & 0.513±0.040          & 0.127±0.011          & 0.633±0.094          & 0.632±0.044          & 0.317±0.030          & 0.079±0.017          \\
LWGP {[}\textit{TCYB}, 2018{]}           & 0.226±0.019          & 0.417±0.046          & 0.590±0.019          & 0.121±0.011          & 0.971±0.010          & 0.658±0.048          & 0.345±0.020          & 0.084±0.015          \\
RSEC-H {[}\textit{TKDD}, 2019{]}         & 0.180±0.027          & 0.477±0.110          & 0.475±0.082          & 0.094±0.030          & 0.973±0.048          & 0.445±0.057          & 0.360±0.046          & 0.060±0.025          \\
RSEC-Z {[}\textit{TKDD}, 2019{]}         & 0.201±0.032          & 0.397±0.149          & 0.483±0.076          & 0.100±0.025          & 0.892±0.080          & 0.485±0.041          & 0.325±0.050          & 0.053±0.023          \\
DREC {[}\textit{Neurocomputing}, 2019{]} & 0.257±0.026          & 0.679±0.079          & 0.544±0.008          & 0.126±0.014          & 0.897±0.076          & 0.703±0.042          & 0.367±0.038    & 0.093±0.014    \\
SPCE {[}\textit{TNNLS}, 2021{]}          & 0.317±0.106          & {\ul 0.737±0.057}    & 0.475±0.127          & 0.113±0.037          & 0.965±0.054          & 0.578±0.074          & {\ul 0.369±0.039}          & 0.088±0.021          \\
ECPCS-MC {[}\textit{TSMC-S}, 2021{]}     & 0.271±0.019          & 0.520±0.039          & 0.546±0.036          & 0.113±0.011          & 0.928±0.068          & 0.611±0.035          & 0.328±0.035          & 0.067±0.005          \\
ECPCS-HC {[}\textit{TSMC-S}, 2021{]}     & 0.379±0.025    & 0.731±0.020          & 0.583±0.076          & 0.139±0.017    & {\ul 0.981±0.033}    & 0.633±0.052          & 0.317±0.039          & 0.067±0.015          \\
TRCE {[}\textit{AAAI}, 2021{]}           & 0.237±0.028          & 0.643±0.128          & 0.588±0.038          & 0.095±0.022          & 0.946±0.085          & 0.621±0.048          & 0.321±0.043          & 0.059±0.017          \\ \midrule
LWEA {[}\textit{TCYB}, 2018{]}           & 0.367±0.033          & 0.430±0.030          & 0.558±0.060          & 0.130±0.016          & 0.928±0.096          & 0.705±0.045    & 0.348±0.038          & 0.084±0.018          \\
Proposed ($\alpha=0.8, \lambda=0.4$)                 & {\ul 0.401±0.054}    & 0.487±0.104          & {\ul 0.644±0.038}    & {\ul 0.166±0.025}    & 0.969±0.065          & {\ul 0.718±0.052}    & 0.365±0.049          & {\ul 0.103±0.018}    \\
Proposed                       & \textbf{0.525±0.075} & \textbf{0.754±0.018} & \textbf{0.657±0.024} & \textbf{0.179±0.017} & \textbf{0.989±0.004} & \textbf{0.723±0.052} & \textbf{0.371±0.056} & \textbf{0.105±0.022} \\ \bottomrule
\end{tabular}}
\end{table*}

\subsubsection{Sensitivity of hyper-parameters}
There are two hyper-parameters $\alpha$ and $\lambda$ in our model, where $\alpha$ decides to what extent we can accept an entry in $\widetilde{\mathbf{A}}$ as a high-confidence element, and $\lambda$ balances the importance of the error term. We investigated their influence on the proposed model in Figs.\,\ref{varyalp} and \ref{varylam}. 
\begin{itemize}
\item First, it can be seen from Fig.\,\ref{varyalp} that the optimal performance is achieved when $\alpha$ is $0.7$ on Caltech20, $0.75$ on Ecoli and $0.8$ on all other datasets. Moreover, an apparent improvement of our method over the baseline LWEA usually occurs when $\alpha\in$ $\{0.7, 0.75, 0.8\}$ since a bigger $\alpha$ will largely reduce the amount of the highly-reliable information, while a smaller $\alpha$ will degrade the quality of HC matrix.
\item Second, from Fig.\,\ref{varylam}, we can observe that when $\lambda$ is 
too large (e.g., $\lambda\geq 100$), the performance of the proposed model will be close to that of LWEA, as with a large $\lambda$, the self-enhanced CA matrix will be close to the original one. 
$\lambda$ should also not be too small, as with a small $\lambda$, the majority of elements of the original CA matrix will be regarded as noise and be removed. 
Moreover,  our model significantly surpasses the baseline on a wide range of $\lambda$, i.e., $\lambda\in[0.01, 1]$ on most of the datasets, proving its robustness. 
\end{itemize}
As a summary, we suggest setting $\alpha=0.8$ and $\lambda=0.4$ for our model.

\subsubsection{Influence of ADMM parameters}
The adopted optimization method ADMM introduces two parameters $\gamma_1$ and $\gamma_2$. In all the experiments, we fixed them as $\gamma_1=\gamma_2=1$. Here, we studied their influence on the proposed method.
Table\,\ref{tab:ADMM_para} shows when varying the values of $\gamma_1$ and $\gamma_2$ from a wide range $[0.1,10]$, the clustering performance barely changes. Besides, in that range, the converged variables of the proposed model also do not change dramatically. Therefore, we can conclude that ADMM is quite robust to $\gamma_1$ and $\gamma_2$, and setting $\gamma_1=\gamma_2=1$ in all the experiments is a reasonable choice.

\subsection{Clustering Performance Comparison}
\begin{table*}[t]
\centering
\caption{Clustering performances of different algorithms on two large-scale datasets measured by ARI, NMI and F-score.}
\label{tab:large_scale}
\begin{tabular}{@{}l|ccc|ccc@{}}
\toprule
\multicolumn{1}{c|}{Dataset}   & \multicolumn{3}{c|}{ISOLET}                                        & \multicolumn{3}{c}{USPS}                                           \\ \midrule
\multicolumn{1}{c|}{Method}    & ARI                  & NMI                  & F-score              & ARI                  & NMI                  & F-score              \\ \midrule
Base clusterings (average)       & 0.404±0.074          & 0.697±0.059          & 0.426±0.070          & 0.215±0.066          & 0.535±0.055          & 0.249±0.079          \\
Base clusterings (best)       & 0.516                & 0.739                & 0.537                & 0.339                & 0.578                & 0.399                \\
PTA-AL {[}\textit{TKDE}, 2016{]}         & 0.522±0.012          & 0.752±0.004          & 0.542±0.012          & 0.535±0.043          & 0.662±0.026          & 0.586±0.038          \\
PTA-CL {[}\textit{TKDE}, 2016{]}         & 0.527±0.008          & 0.752±0.005          & 0.546±0.008          & 0.384±0.040          & 0.557±0.032          & 0.450±0.036          \\
PTGP {[}\textit{TKDE}, 2016{]}           & 0.526±0.022          & 0.752±0.009          & 0.546±0.021          & 0.531±0.050          & 0.649±0.029          & 0.580±0.044          \\
LWGP {[}\textit{TCYB}, 2018{]}           & 0.514±0.017          & 0.753±0.006          & 0.535±0.016          & 0.477±0.035          & 0.643±0.023          & 0.535±0.031          \\
DREC {[}\textit{Neurocomputing}, 2019{]} & 0.531±0.011          & 0.755±0.006          & 0.550±0.011          & \textbf{0.546±0.045} & {\ul 0.667±0.028}    & \textbf{0.593±0.040} \\
SPCE {[}\textit{TNNLS}, 2021{]}          & 0.435±0.055          & 0.759±0.018          & 0.465±0.050          & 0.402±0.064          & 0.605±0.032          & 0.473±0.050          \\
ECPCS-MC {[}\textit{TSMC-S}, 2021{]}     & 0.513±0.017          & 0.751±0.016          & 0.533±0.004          & 0.472±0.012          & 0.627±0.011          & 0.530±0.012          \\
ECPCS-HC {[}\textit{TSMC-S}, 2021{]}     & 0.522±0.021          & 0.757±0.020          & 0.543±0.008          & 0.454±0.024          & 0.623±0.019          & 0.520±0.017          \\
TRCE {[}\textit{AAAI}, 2021{]}           & 0.503±0.007          & 0.754±0.025          & 0.524±0.024          & 0.471±0.021          & 0.650±0.023          & 0.534±0.019          \\ \midrule
LWEA {[}\textit{TCYB}, 2018{]}           & {\ul 0.553±0.016}    & 0.764±0.008          & 0.573±0.015          & 0.525±0.033          & 0.663±0.019          & 0.578±0.028          \\
Proposed ($\alpha=0.8, \lambda=0.4$)                 & {\ul 0.553±0.020}    & {\ul 0.775±0.008}    & {\ul 0.574±0.019}    & 0.519±0.035          & 0.663±0.019          & 0.574±0.029          \\
Proposed                       & \textbf{0.562±0.014} & \textbf{0.774±0.007} & \textbf{0.582±0.013} & {\ul 0.540±0.040}    & \textbf{0.672±0.019} & {\ul 0.591±0.034}    \\ \bottomrule
\end{tabular}
\end{table*}

\begin{figure}[t]
\centering
\subfigure[LWCA]{
\begin{minipage}[t]{0.47\linewidth}
\centering
\includegraphics[width=1.64in]{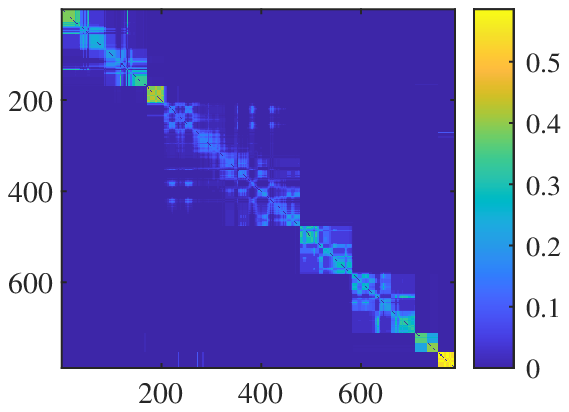}
\end{minipage}}
\subfigure[Proposed Method]{
\begin{minipage}[t]{0.47\linewidth}
\centering
\includegraphics[width=1.64in]{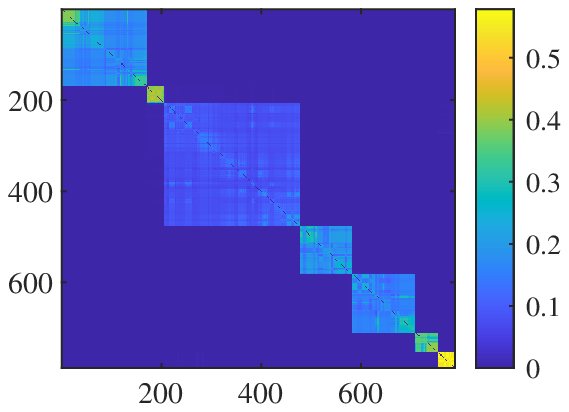}
\end{minipage}}
\subfigure[Difference]{
\label{diff}
\begin{minipage}[t]{0.47\linewidth}
\centering
\includegraphics[width=1.64in]{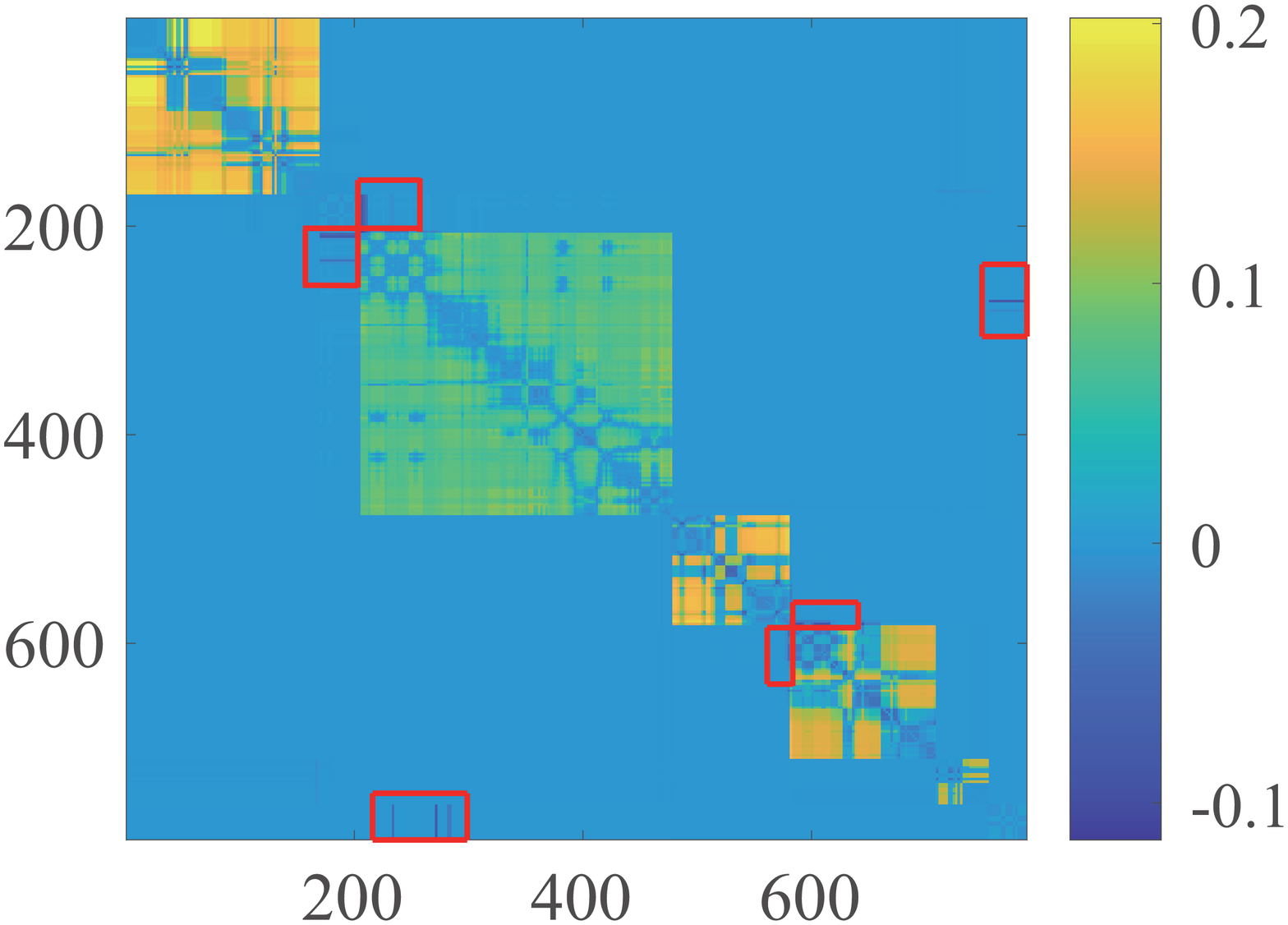}
\end{minipage}}
\subfigure[Ground Truth]{
\begin{minipage}[t]{0.47\linewidth}
\centering
\includegraphics[width=1.64in]{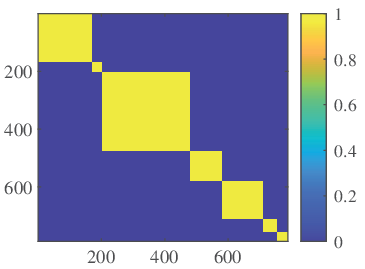}
\end{minipage}}
\centering
\caption{Comparison of the LWCA, the proposed CA matrix, their difference and the ideal one produced by ground truth labels.}
\label{visual}
\end{figure}

\subsubsection{Results Comparison}
Table\,\ref{tab:ari} shows the average ARI as well as the standard deviation of the compared and proposed algorithms on the eight modest-scale benchmark datasets. We provide two groups of results with our algorithm: Proposed and Proposed ($\alpha=0.8, \lambda=0.4$). The former demonstrates the best-tuned results of the proposed algorithm w.r.t. ARI and the latter corresponds to the proposed algorithm with fixing hyper-parameters $\alpha=0.8$ and $\lambda=0.4$. Note that for the compared methods, their hyper-parameters were carefully tuned, and the best performance was reported.  Additionally, the best and average performance of the base clusterings from the candidate base clustering pool were also included.
The clustering performances with other metrics (NMI, F-Score, Accuracy, Purity) are shown in the supplementary file. From those tables, we have the following observations.
\begin{itemize}
    \item First, nearly all the ensemble clustering methods perform better than the average performance of the base clusterings with different metrics, which indicates that leveraging multiple base clusterings is useful. But many compared methods cannot exceed the best base clustering on most datasets, e.g., PTA-CL and RSEC-Z, suggesting their limitations. Differently, the proposed method can significantly outperform both the average performance of the base clusterings and the best base clustering in the candidate base clustering pool on most datasets, which proves that the proposed method can really exploit the useful information from the base clusterings and improve the clustering performance.
    \item  Second, the proposed method takes the CA matrix of LWEA as input, and it can largely improve the clustering performance of LWEA on all the datasets, e.g., on Ecoli, our method improves the ARI from $0.430$ to $0.754$, NMI from $0.615$ to $0.716$ and F-score from $0.544$ to $0.823$. This is a straightforward proof that our method can improve the quality of the input CA matrix.
    \item Third, the proposed method outperforms all the compared methods on all eight datasets w.r.t. ARI and F-score. For example, on Caltech20, the proposed method improves ARI by about $40\%$ compared with the best comparison. In terms of NMI, our method achieves the best performance on 6 out of 8 cases, while is slightly inferior to SPCE on UMIST and SPF. Taking all the metrics into consideration, we can conclude that the proposed model ranks first in general.
    \item Finally, from the results of the Proposed ($\alpha=0.8, \lambda=0.4$), it can be found that without tuning hyper-parameters, the proposed method can still achieve the second-best performance on average (slightly inferior to the tuned one). This phenomenon proves that our method is very practical in real-world applications.
\end{itemize}

\begin{table}[t]
\centering
\caption{Illustration of the clustering improvement on EAC w.r.t. ARI, NMI and F-score.}
\label{withCA}
\resizebox{\linewidth}{!}{
\begin{tabular}{@{}c|cc|cc|cc@{}}
\toprule
Metric      & \multicolumn{2}{c|}{ARI} & \multicolumn{2}{c|}{NMI} & \multicolumn{2}{c}{F-score} \\ \midrule
Method      & EAC    & Proposed        & EAC    & Proposed        & EAC      & Proposed         \\ \midrule
Caltech20   & 0.313  & \textbf{0.490}  & 0.467  & \textbf{0.495}  & 0.386    & \textbf{0.573}   \\
Ecoli       & 0.548  & \textbf{0.753}  & 0.651  & \textbf{0.717}  & 0.649    & \textbf{0.823}   \\
LS          & 0.467  & \textbf{0.638}  & 0.563  & \textbf{0.675}  & 0.575    & \textbf{0.714}   \\
FCT         & 0.121  & \textbf{0.146}  & 0.210  & \textbf{0.241}  & 0.268    & \textbf{0.305}   \\
Aggregation & 0.896  & \textbf{0.989}  & 0.942  & \textbf{0.984}  & 0.916    & \textbf{0.991}   \\
Texture     & 0.575  & \textbf{0.673}  & 0.713  & \textbf{0.802}  & 0.617    & \textbf{0.707}   \\
UMIST       & 0.333  & \textbf{0.368}  & 0.651  & \textbf{0.690}  & 0.372    & \textbf{0.407}   \\
SPF         & 0.057  & \textbf{0.086}  & 0.123  & \textbf{0.191}  & 0.271    & \textbf{0.358}   \\ \bottomrule
\end{tabular}}
\end{table}

\begin{table}[t]
\centering
\caption{Illustration of the clustering improvement on PTA-AL w.r.t. ARI, NMI and F-score.}
\label{withPTA}
\resizebox{\linewidth}{!}{
\begin{tabular}{@{}c|cc|cc|cc@{}}
\toprule
Metric      & \multicolumn{2}{c|}{ARI} & \multicolumn{2}{c|}{NMI} & \multicolumn{2}{c}{F-score} \\ \midrule
Method      & PTA    & Proposed        & PTA    & Proposed        & PTA      & Proposed         \\ \midrule
Caltech20   & 0.356  & \textbf{0.439}  & 0.440  & \textbf{0.475}  & 0.436    & \textbf{0.537}   \\
Ecoli       & 0.340  & \textbf{0.738}  & 0.561  & \textbf{0.700}  & 0.460    & \textbf{0.811}   \\
LS          & 0.593  & \textbf{0.664}  & 0.635  & \textbf{0.687}  & 0.673    & \textbf{0.735}   \\
FCT         & 0.136  & \textbf{0.190}  & 0.242  & \textbf{0.277}  & 0.271    & \textbf{0.335}   \\
Aggregation & 0.600  & \textbf{0.896}  & 0.786  & \textbf{0.932}  & 0.681    & \textbf{0.920}   \\
Texture     & 0.658  & \textbf{0.702}  & 0.772  & \textbf{0.800}  & 0.691    & \textbf{0.730}   \\
UMIST       & 0.323  & \textbf{0.370}  & 0.633  & \textbf{0.687}  & 0.361    & \textbf{0.411}   \\
SPF         & 0.081  & \textbf{0.098}  & 0.152  & \textbf{0.188}  & 0.261    & \textbf{0.337}   \\ \bottomrule
\end{tabular}}
\end{table}

\begin{figure*}[t]
\centering
\subfigure{
\begin{minipage}[t]{0.23\linewidth}
\centering
\includegraphics[width=1.6in]{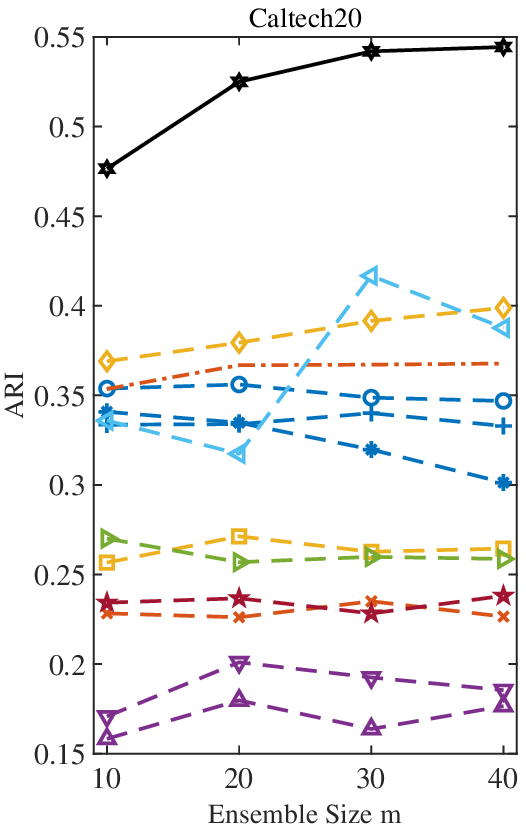}
\end{minipage}}
\subfigure{
\begin{minipage}[t]{0.23\linewidth}
\centering
\includegraphics[width=1.6in]{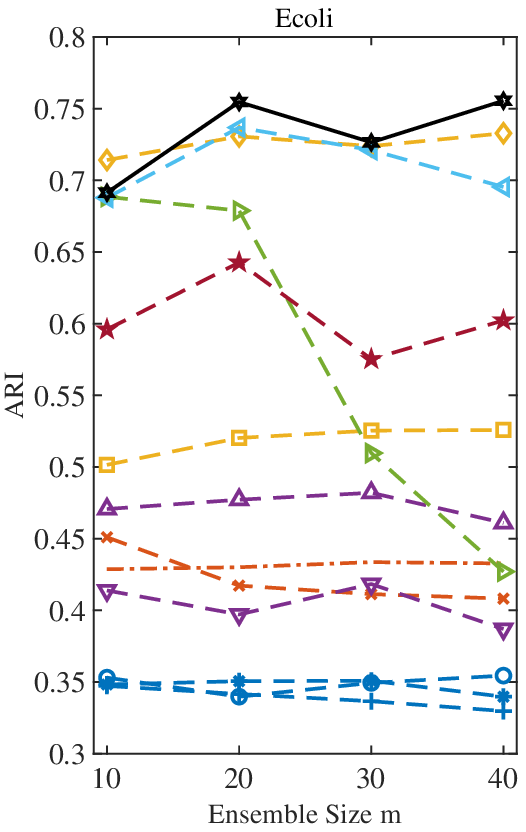}
\end{minipage}}
\subfigure{
\begin{minipage}[t]{0.23\linewidth}
\centering
\includegraphics[width=1.6in]{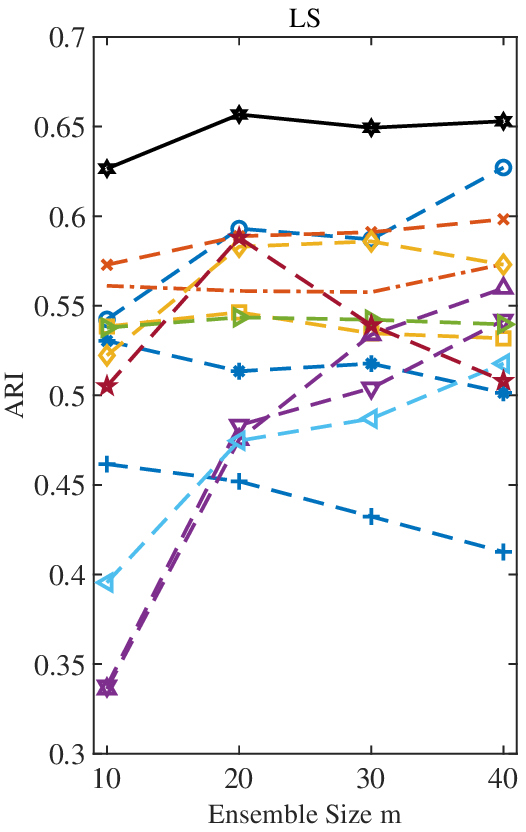}
\end{minipage}}
\subfigure{
\begin{minipage}[t]{0.23\linewidth}
\centering
\includegraphics[width=1.6in]{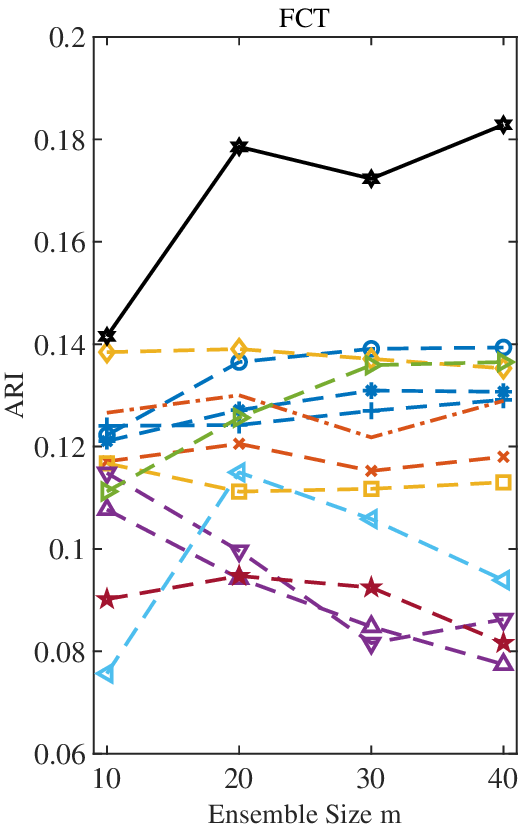}
\end{minipage}}
\subfigure{
\begin{minipage}[t]{0.23\linewidth}
\centering
\includegraphics[width=1.6in]{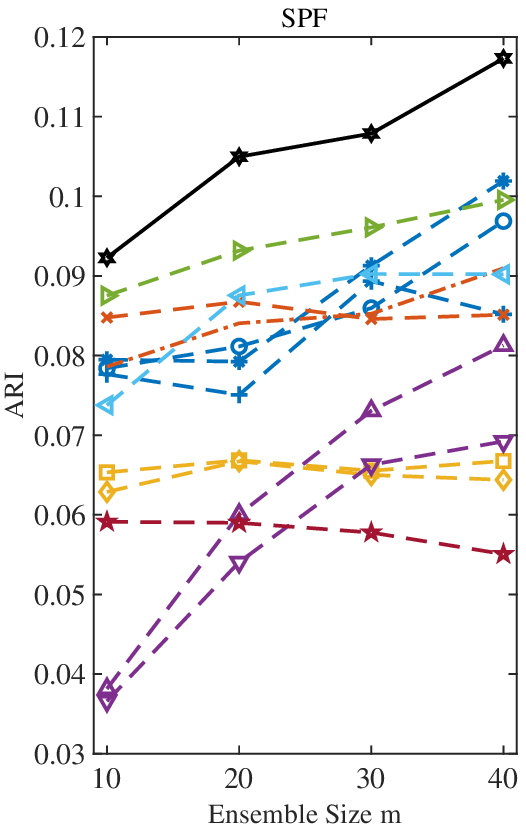}
\end{minipage}}
\subfigure{
\begin{minipage}[t]{0.23\linewidth}
\centering
\includegraphics[width=1.6in]{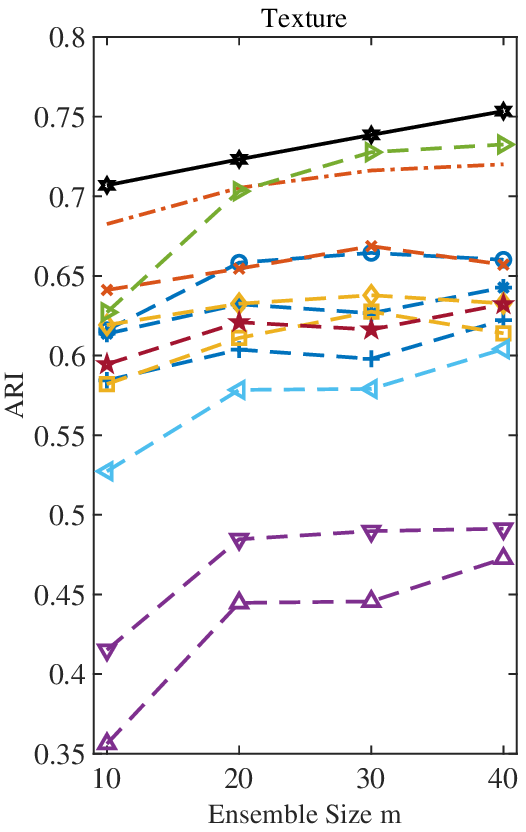}
\end{minipage}}
\subfigure{
\begin{minipage}[t]{0.23\linewidth}
\centering
\includegraphics[width=1.6in]{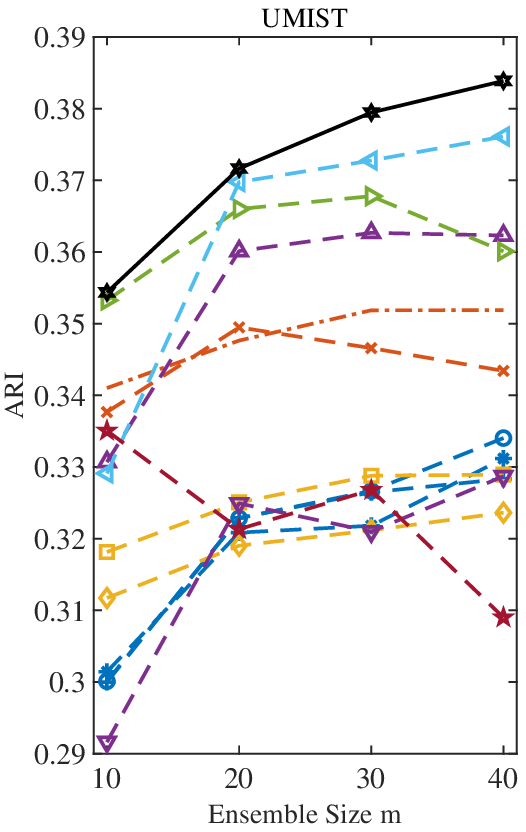}
\end{minipage}}
\subfigure{
\begin{minipage}[t]{0.23\linewidth}
\centering
\includegraphics[width=1.6in]{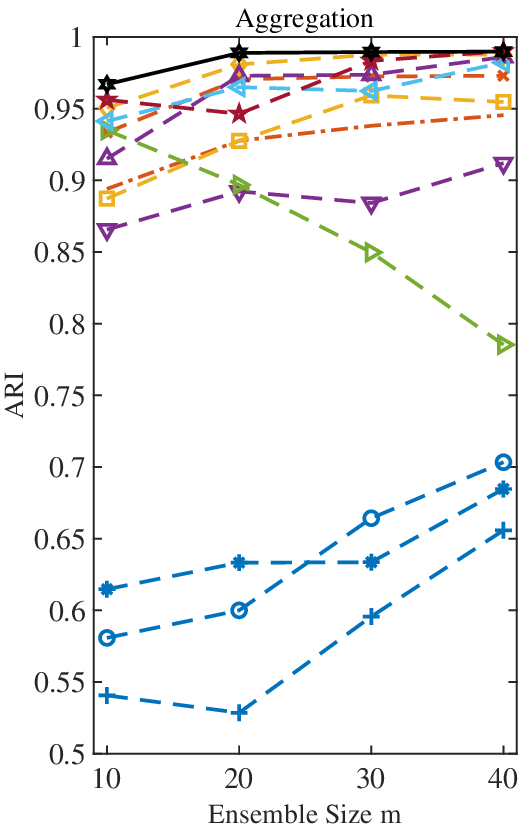}
\end{minipage}}
\subfigure{
\begin{minipage}[t]{0.96\linewidth}
\centering
\includegraphics[width=6.8in]{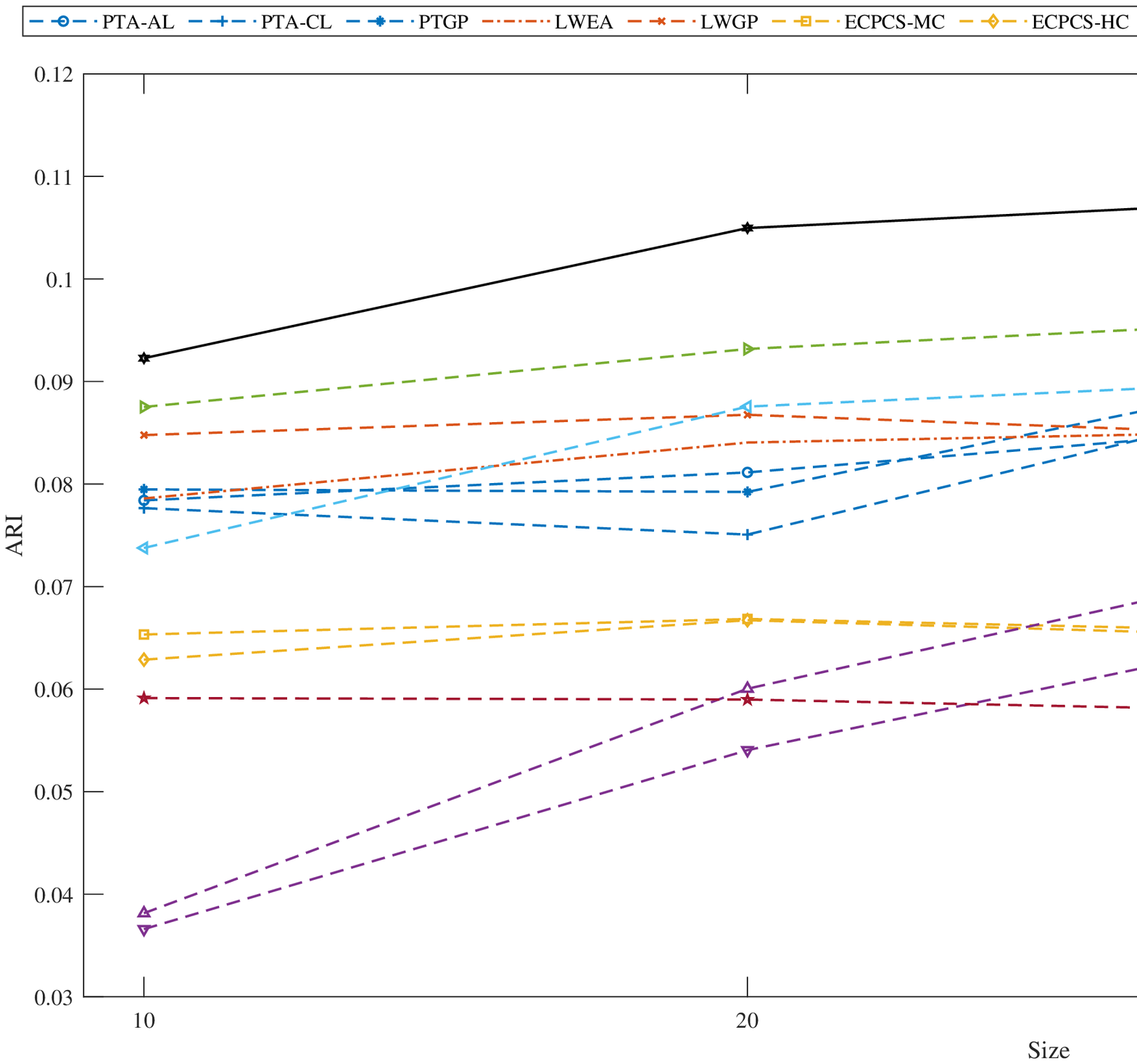}
\end{minipage}}
\centering
\caption{Clustering performances of different algorithms by varying ensemble size w.r.t. ARI.}
\label{size}
\end{figure*}

\subsubsection{Results on Large-scale Datasets}
In Table\,\ref{tab:large_scale}, we also tested the performance of the proposed algorithm on two large-scale datasets, i.e., ISOLET and USPS. We can observe that on large-scale datasets, the proposed model also outperforms the state-of-the-art ensemble clustering methods in most cases with different metrics. 

\subsubsection{Visualization}
Fig.\,\ref{visual} visually compares the LWCA matrix, the CA matrix of our method, their difference, and the ground truth CA matrix on the Aggregation dataset. We can see that the proposed method can considerably build more strong and more reliable links than LWCA, which is much closer to the ideal one. Moreover, it can also weaken the wrong entries of the LWCA matrix (please zoom in on the highlighted regions in Fig.\,\ref{diff}). This further explains why our method can significantly improve the clustering performance compared with LWEA.

\begin{table*}[t]
\centering
\caption{Ablation study on the proposed method w.r.t. ARI, NMI and F-score.}
\label{tab:ablation}
\begin{tabular}{@{}cccccccccc@{}}
\toprule
Metric                  & Algorithm                                                              & Caltech20      & Ecoli          & LS             & FCT            & Aggregation    & Texture        & UMIST          & SPF            \\ \midrule
\multirow{3}{*}{ARI}     & w/o $\operatorname{tr}(\mathbf{C}^{\top} \boldsymbol{\Phi} \mathbf{C})$ & 0.367          & 0.430          & 0.558          & 0.130          & 0.928          & 0.705          & 0.348          & 0.084          \\
                         & w/o $\Psi_\Omega(\mathbf{E})=\mathbf{0}$                                                        & 0.515          & 0.699          & 0.646          & 0.174          & 0.976          & \textbf{0.732} & \textbf{0.373} & 0.102          \\
                         & Proposed                                                               & \textbf{0.525} & \textbf{0.754} & \textbf{0.657} & \textbf{0.179} & \textbf{0.989} & 0.723          & 0.371          & \textbf{0.105} \\ \midrule
\multirow{3}{*}{NMI}     & w/o $\operatorname{tr}(\mathbf{C}^{\top} \boldsymbol{\Phi} \mathbf{C})$ & 0.480          & 0.615          & 0.630          & 0.236          & 0.958          & 0.794          & 0.664          & 0.156          \\
                         & w/o $\Psi_\Omega(\mathbf{E})=\mathbf{0}$                                                        & \textbf{0.497} & 0.690          & 0.677          & 0.273          & 0.978          & \textbf{0.813} & 0.685          & 0.197          \\
                         & Proposed                                                               & \textbf{0.497} & \textbf{0.716} & \textbf{0.685} & \textbf{0.276} & \textbf{0.984} & 0.811          & \textbf{0.692} & \textbf{0.198} \\ \midrule
\multirow{3}{*}{F-score} & w/o $\operatorname{tr}(\mathbf{C}^{\top} \boldsymbol{\Phi} \mathbf{C})$ & 0.441          & 0.544          & 0.649          & 0.260          & 0.941          & 0.733          & 0.386          & 0.280          \\
                         & w/o $\Psi_\Omega(\mathbf{E})=\mathbf{0}$                                                        & 0.587          & 0.786          & 0.720          & 0.311          & 0.981          & \textbf{0.757} & \textbf{0.413} & 0.344          \\
                         & Proposed                                                               & \textbf{0.603} & \textbf{0.823} & \textbf{0.729} & \textbf{0.322} & \textbf{0.991} & 0.750          & 0.411          & \textbf{0.348} \\ \bottomrule
\end{tabular}
\end{table*}

\subsection{Adaptation to Diverse CAs}
\label{adapt_CA}
In this subsection, we check whether the proposed approach can adapt to other CA matrices except for LWCA. Here we used the CA matrix from EAC \cite{fred2005combining} and the similarity matrix PTS from PTA \cite{huang2015robust} as the input, and assessed the performance of our method on them, respectively. The associated results are listed in Tables\,\ref{withCA} and \ref{withPTA}, where the proposed model consistently promotes the clustering performances of original algorithms to a great extent on the eight modest-scale datasets with different metrics. To be specific, when compared with EAC, the average improvements of our method are around $30\%$ in ARI, $15\%$ in NMI and $20\%$ in F-score. Especially, our method has pushed these three metrics nearly up to 1 on the Aggregation dataset.
The improvements on PTA are $35\%$, $15\%$, $30\%$ on average, and $120\%$, $25\%$, $75\%$ singly on the Ecoli dataset w.r.t. ARI, NMI and F-score, which is remarkable.
Note that after the self-enhancement procedure by our method, when referring to Tables\,\ref{tab:ari} and the results of NMI and F-score in the supplementary file, EAC and PTA-AL will attain the highest ARI/NMI/F-score on 6/5/7 and 6/3/6 datasets over all the compared state-of-the-art algorithms respectively, while these numbers are only 0/0/0 and 1/0/1 corresponding to their original algorithms.
Therefore, we can conclude that the proposed method can enhance diverse CA matrices significantly, illustrating its robustness.

\begin{table*}[t]
\centering
\caption{Execution time of algorithms involving an iteration process on eight modest-scale datasets. We highlight the fastest and the second fastest ones by bold and underline. Execution time of the baseline algorithm---LWEA is added to show how much the proposed method paid to improve the performance.}
\label{time}
\begin{tabular}{@{}ccccccc|c@{}}
\toprule
Time(s)     & RSEC-H \cite{tao2019robust} & RSEC-Z \cite{tao2019robust} & DREC  \cite{zhou2019ensemble}          & SPCE \cite{zhou2020self} & TRCE \cite{zhou2021tri} & Proposed        & Baseline---LWEA \cite{huang2017locally} \\ \midrule
Caltech20   & 700    & 707    & \textbf{2.16}  & 39.4  & 31.5  & {\ul 19.5}      & 0.332 \\
Ecoli       & 2.91   & 2.94   & {\ul 0.130}    & 0.273 & 0.491 & \textbf{0.0862} & 0.061 \\
LS          & 14006  & 14107  & \textbf{21.2}  & 1068  & 354   & {\ul 277}       & 1.85  \\
FCT         & 3633   & 3663   & \textbf{4.24}  & 244   & 131   & {\ul 60.3}      & 0.751 \\
Aggregation & 17.0   & 17.1   & \textbf{0.167} & 2.53  & 2.22  & {\ul 1.17}      & 0.095 \\
Texture     & 8207   & 8271   & \textbf{6.14}  & 409   & 262   & {\ul 164}       & 1.19 \\
UMIST       & 8.41   & 8.48   & {\ul 0.309}    & 2.50  & 2.79  & \textbf{0.292}  & 0.119 \\
SPF         & 285    & 288    & \textbf{0.250} & 39.5  & 17.1  & {\ul 12.5}      & 0.230 \\ \bottomrule
\end{tabular}
\end{table*}

\subsection{Influence of Ensemble Size}
\label{ensemble_size}
Fig.\,\ref{size} evaluates the clustering performance w.r.t. ARI of the proposed method and all the compared methods with different numbers of base clusterings as input. From Fig.\,\ref{size} we can observe that with a bigger ensemble size, most methods generally perform better, which is consistent with the basic idea of ensemble clustering that combining a set of clustering results can generate a better one. It is noticeable that on 5 out of 8 datasets, our method keeps a strictly monotonic increasing ARI when the ensemble size grows. In contrast, some of the compared methods sometimes experience an opposite trend such as DREC and TRCE.
Moreover, it is evident that the proposed method outperforms all the compared methods with different ensemble sizes except case $m=10$ on Ecoli, being the most robust model among all comparisons. On datasets Caltech20, LS and FCT, the performance of our method is far ahead of compared methods, and our method even performs better with $10$ input base clusterings than all the compared methods though they take $40$ base clusterings as input.

\subsection{Ablation Study}
\label{exp_ablation}
In this section, we study the necessity of the two terms involved in our method. First, we dropped the Laplacian regularization term, then the proposed model degenerates into Eq.\,(\ref{initial}), and we named this case as ``w/o $\operatorname{tr}(\mathbf{C}^{\top} \boldsymbol{\Phi} \mathbf{C})$".
Second, we removed the matrix-completion constraint $\Psi_\Omega(\mathbf{E})=\mathbf{0}$, naming it as ``w/o $\Psi_\Omega(\mathbf{E})=\mathbf{0}$".

The clustering performance without those terms is shown in Table\,\ref{tab:ablation}. It is obvious that ``w/o $\operatorname{tr}(\mathbf{C}^{\top} \boldsymbol{\Phi} \mathbf{C})$" always receives the worst scores, indicating that Laplacian regularization plays a leading role in propagating high-confidence information. Additionally, the average performance of ``w/o $\Psi_\Omega(\mathbf{E})=\mathbf{0}$" exceeds that of the proposed method slightly on Texture and UMIST, while falls behind it on the other six datasets distinctly. Thus, it can be concluded that retaining those crucial HC values also contributes to enhancing the co-association matrix.

\subsection{Execution Time}
In Table\,\ref{time}, we count the average running time of our method with five compared methods that also involve an iteration optimization process, i.e., RSEC-H, RSEC-Z, DREC, SPCE and TRCE, on the eight modest-scale datasets. We also show the running time of LWEA as a reference. Note that LWEA does not need iterative optimization (can be computed analytically), which runs faster than other methods. Table\,\ref{time} suggests that the proposed method is more efficient than RSEC-H, RSEC-Z, SPCE and TRCE due to the convexity of the proposed model and the low computational complexity of the proposed optimization algorithm. Besides, our method is more time-saving than DREC on the Ecoli and UMIST datasets, but runs slower than DREC when the scale of the dataset gets larger, mainly because DREC uses the microcluster-based representation that reduces the scale of the problem. Considering that our method significantly outperforms DREC in clustering performance, a little sacrifice in running time is acceptable.

\section{Conclusion}
\label{sec5}
In this paper, we have presented a novel CA-matrix self-enhancement model for ensemble clustering. Without any extra information, our method can promote a CA matrix by exploiting the highly-reliable information and denoising error connections simultaneously. We formulated this model as a well-defined convex optimization problem and proposed an ADMM-based algorithm to solve it with convergence and global optimum theoretically guaranteed.
Extensive experimental comparisons validate that the proposed model outperforms the state-of-the-art methods significantly in clustering performance. Moreover, it is robust to hyper-parameters, can adapt to diverse CA matrices, produces better performance with more base clusterings, and is more efficient than many other approaches.

This paper only uses the high-value elements in the CA matrix as the highly reliable information, however, the low-value elements in the CA matrix are also reliable, which indicates that the two samples do not belong to the same cluster with a high probability. In the future, we will investigate how to exploit such information.

\bibliography{reference}{}
\bibliographystyle{IEEEtran}

\begin{IEEEbiography}[{\includegraphics[width=1in,height=1.25in,clip,keepaspectratio]{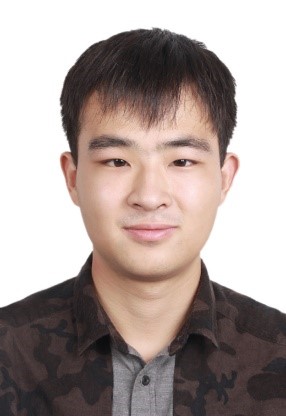}}]{Yuheng Jia}
received the B.S. degree in automation and the M.S. degree in control theory and engineering from Zhengzhou University, Zhengzhou, China, in 2012 and 2015, respectively, and the Ph.D. degree in computer science from the City University of Hong Kong, SAR, China, in 2019.

He is currently an associate professor with the School of Computer Science and Engineering, Southeast University, China. His research interests broadly include topics in machine learning and data representation, such as semi-supervised learning, high-dimensional data modeling and analysis, low-rank tensor/matrix approximation and factorization.
\end{IEEEbiography}

\begin{IEEEbiography}[{\includegraphics[width=1in,height=1.25in,clip,keepaspectratio]{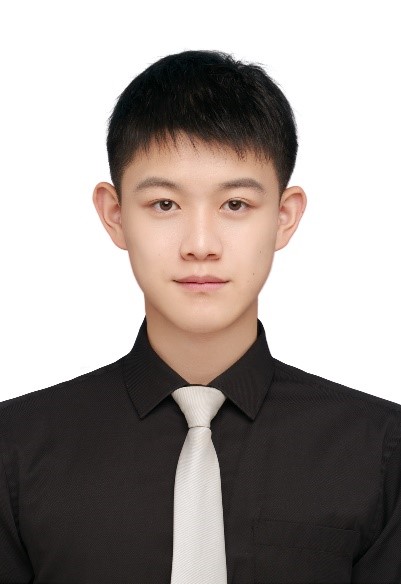}}]{Sirui Tao}
is currently a senior student at the School of Automation, Southeast University, and is admitted to study for an M.Sc. degree in data science at the Academy for Advanced Interdisciplinary Studies, Peking University without entrance examination. His current research interests lie in machine learning and high-performance computing.
\end{IEEEbiography}

\begin{IEEEbiography}[{\includegraphics[width=1in,height=1.25in,clip,keepaspectratio]{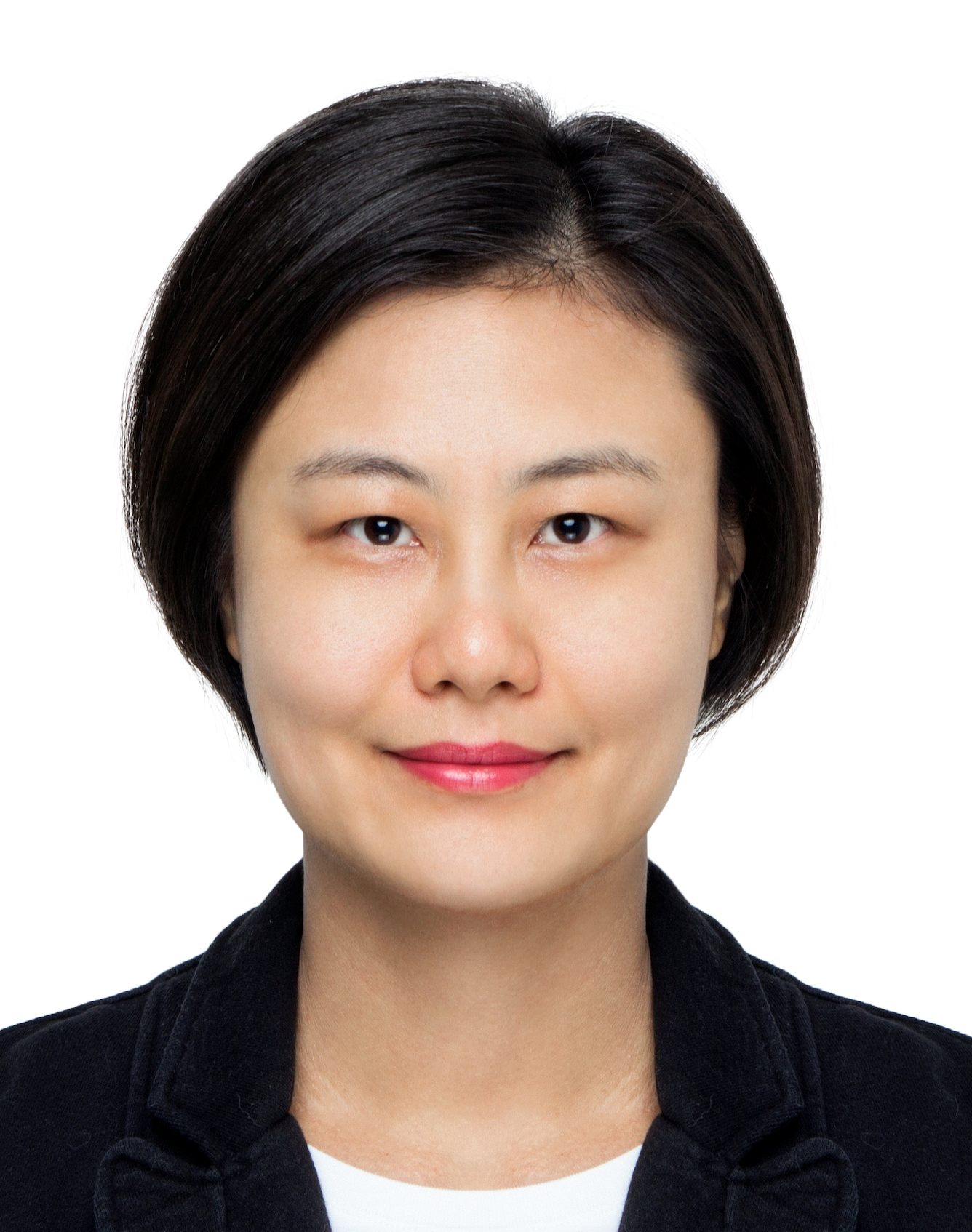}}]{Ran Wang}
(S'09-M'14) received her B.Eng. degree in computer science from the College of Information Science and Technology, Beijing Forestry University, Beijing, China, in 2009, and the Ph.D. degree from the Department of Computer Science, City University of Hong Kong, Hong Kong, in 2014. From 2014 to 2016, she was a Postdoctoral Researcher at the Department of Computer Science, City University of Hong Kong. She is currently an Associate Professor at the College of Mathematics and Statistics, Shenzhen University, China. Her current research interests include pattern recognition, machine learning, fuzzy sets and fuzzy logic, and their related applications.
\end{IEEEbiography}

\begin{IEEEbiography}[{\includegraphics[width=1in,height=1.25in,clip,keepaspectratio]{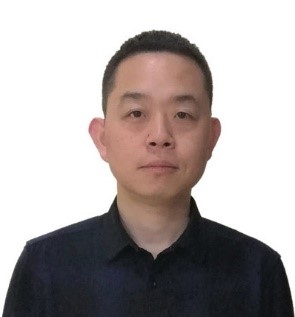}}]{Yongheng Wang}
received the Ph.D degree in computer science and technology from National University of Defense Technology, Changsha, China, in 2006.

He is currently a research specialist in the Research Center for Data Mining and Knowledge Discovery, Zhejiang Lab. His research interests include Big data analytics, machine learning and intelligent decision making. His current research is on intelligent interactive data analysis and simulation-based intelligent decision making in the economic field.
\end{IEEEbiography}

\end{document}